\documentclass{article} 
\usepackage{iclr2027_conference}
\iclrfinalcopy

\usepackage{amsmath,amsfonts,bm}

\def\eqref#1{equation~\ref{#1}}

\def\1{\bm{1}}

\DeclareMathAlphabet{\mathsfit}{\encodingdefault}{\sfdefault}{m}{sl}
\SetMathAlphabet{\mathsfit}{bold}{\encodingdefault}{\sfdefault}{bx}{n}

\newcommand{\KL}{D_{\mathrm{KL}}}

\usepackage{hyperref}
\usepackage{multirow}
\usepackage{graphicx}
\usepackage[most]{tcolorbox}
\usepackage{enumitem}
\usepackage{url}
\usepackage{amsmath}
\usepackage{amssymb}
\usepackage{bm}
\usepackage{algorithm}
\usepackage{algpseudocode}
\usepackage{wrapfig}
\usepackage{graphicx}      
\usepackage{booktabs}      
\usepackage{multirow}      
\usepackage{makecell}      
\usepackage{amsmath}       
\usepackage{algorithm}     
\usepackage{algpseudocode} 
\usepackage{colortbl}
\usepackage{arydshln}
\newcommand{\graydashline}{%
  \noalign{\vskip 1pt}%
  \arrayrulecolor{gray!50}\hdashline\arrayrulecolor{black}%
  \noalign{\vskip 1pt}}
\DeclareMathOperator{\JSD}{JSD}
\newcommand{\sg}[1]{\mathrm{sg}\!\left[#1\right]}

\title{MIRROR: From Imitation to Internalization in LLM Personalization}
\author{
Huayi Lai$^{1}$\thanks{Equal contribution.},
Jicheng Yang$^{1}$\footnotemark[1],
Min Yi$^{1}$,
Chong Meng$^{1}$\thanks{Corresponding author.}\\
$^{1}$Baidu Inc.\\
\texttt{\{yangjicheng02,mengchong01\}@baidu.com}
}

\begin{document}
\maketitle

\begin{abstract}
The demand for personalized LLMs is shifting from style imitation toward content quality. We investigate whether self-distillation can bridge this gap in existing fine-tuning paradigm. 
To address this limitation, we introduce MIRROR(Meta-
personalization by Internalizing Reference-Revealed On-policy Reflections),a novel self-distillation framework that shifts LLM personalization from imitation toward preference internalization. 
First, we replace reference-token imitation with reference-revealed on-policy self-distillation, aligning the model’s next-token distributions along its own generation trajectories with those of its reference-conditioned self, thereby internalizing user preferences rather than reproducing reference wording.
Second, we introduce MIRROR-F, a focal plug-in that augments on-policy distributional alignment with selective supervision over informative reference tokens, thereby strengthening content generation while preserving user-specific expression.
Across three personalized generation benchmarks, two model scales, and complementary reference-based and LLM-based evaluations, MIRROR and MIRROR-F achieve leading overall personalization performance and superior text quality, while exhibiting less catastrophic forgetting than SFT-based baselines on three unseen personalized generation tasks.
The gains are consistent across model scales and application scenarios, translating to improved performance in LLM personalization tasks. 
\end{abstract}

\section{Introduction}
\label{Introduction}
Large Language Models (LLMs) have exhibited exceptional capabilities across a wide range of domains~\citep{kimik3,qwen3}, leading to their widespread deployment in real-world applications such as personalized office assistant~\citep{office1,office2} and Private medical diagnosis~\citep{hospital}.
Although various methods enhance the personalization capabilities of general-purpose LLMs through sophisticated harness designs~\citep{harness} and skill injection~\citep{skill}, these approaches largely rely on external mechanisms. From the perspective of the model intelligence, the core challenge remains to strengthen its intrinsic ability to infer latent user states, capture implicit preferences, and perform preference-aware reasoning~\citep{antiharness,antiharness2,antiharness3}.

\begin{figure}[t]
    \centering
    \includegraphics[width=0.95\textwidth]{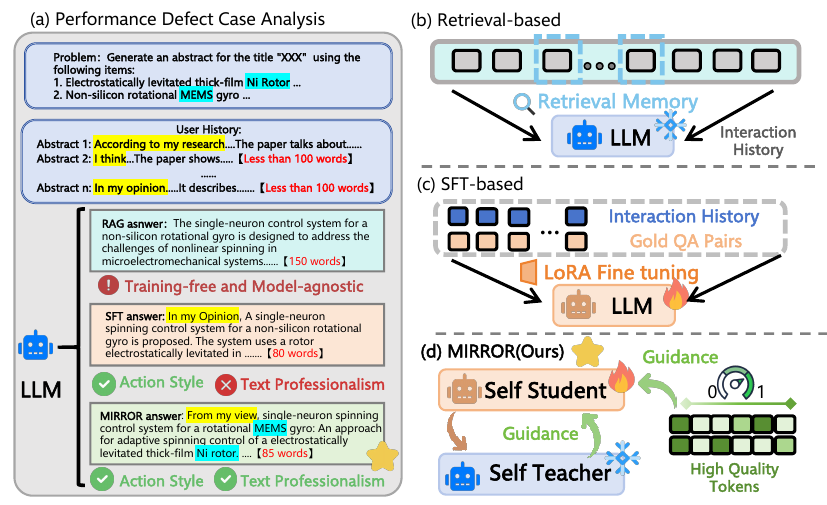}
    \caption{Overview of personalization paradigms. (a) A case study of the quality
    defects of existing methods. (b) Retrieval-based personalization. (c) SFT-based
    personalization. (d) MIRROR, which aligns a self-student with a self-teacher on the
    student's own on-policy trajectories.}
    \label{fig:motivation}
\end{figure}

Mainstream approaches for LLM personalization can be broadly categorized into 3 categories: retrieval-based,fine-tuning-based and user-based methods. First, retrieval-based methods(as shown in Figure~\ref{fig:motivation}.b) incorporate relevant user histories into the input context to elicit preference-aware generation through in-context learning~\citep{rag1,rag2,rag3,icl}. However, their reliance on external contextual information leaves the model's intrinsic preference-understanding capabilities largely unchanged and introduces potential privacy and security risks~\citep{saferag}. 
Second, fine-tuning-based methods(as shown in Figure~\ref{fig:motivation}.c) employ parameter-efficient adaptation with preference-aware token- or sample-level supervision to learn user-specific linguistic patterns~\citep{PerCE,nextquill}. 
While this method is effective at learning personalized styles and sentence structures, they significantly degrades the text professionalism, and can lead to catastrophic forgetting~\citep{forget} and poor out-of-distribution generalization~\citep{ood,pplug}.
Third, user-based methods including user-specific LoRA adapter training~\citep{oppu,betterlora,OPDpersona} ,attention-head pruning~\citep{attention} and steering-vector injection~\citep{perfit}. However, these methods typically require user-specific parameters or dedicated architectures, leading to prohibitive training and storage costs as the user base scales~\citep{nextquill}, which hinders the deployment of personalized LLMs in real world applications.

To date, as shown in Figure~\ref{fig:motivation}.a,existing personalization approaches have been predominantly constrained by a style imitation paradigm, where personalization is primarily modeled as the ability to reproduce a user's linguistic style, expression patterns, and textual distributions. However, this formulation overlooks a fundamental requirement of personalized generation: the generated content must not only reflect a user's expressive preferences but also maintain high standards of factual accuracy, completeness, and practical utility while satisfying the user's underlying needs. Therefore, this paper investigates the following central question: \textbf{How can we enhance the modeling of user-specific expressive styles while maintaining the quality and effectiveness of personalized text generation?}

Recent advances in self-distillation~\citep{OPSD,SDFT} introduce a paradigm beyond demonstration imitation. By leveraging hindsight information, models can internalize preference-relevant behavioral shifts through self-supervision along their own generation trajectories. In LLM personalizaton, user reference information can be viewed as a manifestation of the user's preferences for stylistic sentence structures and their level of expertise, so we can identify user-provided references as a natural form of hindsight information for personalized generation.

Based on this insight, we propose \textbf{MIRROR} (\textbf{M}eta-personalization by \textbf{I}nternalizing \textbf{R}eference-\textbf{R}evealed \textbf{O}n-policy \textbf{R}eflections), a reference-revealed self-distillation framework that shifts personalization from imitation toward internalization. MIRROR (shown in Figure~\ref{fig:motivation}.d) employs a single model as both a self-teacher and a self-student, the teacher accesses the user-provided reference to provide preference-aware guidance, while the student aligns with this guidance along its own on-policy trajectories instead of fitting the reference sequence token by token. Building upon this training paradigm, we further introduce \textbf{MIRROR-F}, a focal personalization framework that selectively concentrates optimization on informative tokens responsible for content quality. This focal mechanism enhances the expertise, completeness, and practical utility of generated content while preserving user-specific expressive styles. Controlled by a continuous focality coefficient. Extensive experiments on three benchmarks demonstrate that MIRROR achieves strong performance across multiple personalized generation tasks and MIRROR-F further achieves the best overall performance under comprehensive personalized generation evaluations, demonstrating superior content expertise, completeness, and textual quality while preserving user-specific expressive preferences.

The main contribution of this work can be summarized as follows:

\begin{itemize}[leftmargin=1.2em,itemsep=1pt,topsep=2pt,parsep=0pt,partopsep=0pt]
    \item We introduce a new perspective by formulating personalized generation as reference-revealed self-distillation, moving beyond surface-level imitation toward user preference internalization.

    \item We propose \textbf{MIRROR}, a self-distillation framework that transfers preference-induced behaviors via on-policy learning, achieving state-of-the-art performance across diverse personalized generation tasks.
    
    \item We further propose \textbf{MIRROR-F}, which selectively enhances informative content generation through focal supervision, improving expertise, completeness, and textual quality while preserving user-specific expressions.
    
    \item Extensive experiments demonstrate that MIRROR and MIRROR-F achieve superior personalization and generation quality across comprehensive personalized generation evaluations.
\end{itemize}

\section{Preliminaries}
\label{Preliminaries}

\paragraph{Personalized text generation.}
We consider the task of personalized text generation, where the goal is to adapt a
language model $\pi_\theta$ so that its outputs match the preferences of an individual
user. Each data sample is a triplet $(x, h, y)$: $x$ is the query specifying the task and
its content, $h$ is the user's historical text that implicitly reflects their preferences,
and $y = (y_1, \dots, y_T)$ is the reference response actually written by the user. The
collection of such samples over all users forms the dataset $\mathcal{D}$, with
$(x, h, y) \in \mathcal{D}$. A non-personalized model produces
$\hat{y} = \pi_\theta(x)$, whereas a personalized model additionally conditions on the
user history, $\hat{y} = \pi_\theta(x, h)$. Our goal is to improve $\pi_\theta$ such that
$\hat{y}$ both reflects the user's expressive preferences and remains accurate and
complete as a response to $x$.

\paragraph{Autoregressive factorization and contexts.}
We write $\pi_\theta(\cdot \mid c, y_{<t}) \in \Delta^{|\mathcal{V}|}$ for the next-token
distribution over the vocabulary $\mathcal{V}$ at step $t$, given a context $c$ and a
prefix $y_{<t}$. Two contexts are used throughout this paper. The
\emph{deployable context} $c_{\mathrm{S}} = (h, x)$ contains only what is genuinely
available at inference time, and the \emph{reference-revealed context}
$c_{\mathrm{T}} = (h, x, y)$ additionally exposes the user-written reference. Since the
reference is unavailable at deployment, a model conditioned on $c_{\mathrm{T}}$ can serve
only as a source of supervision during training, never as a deployed policy. We use
$\sg{\cdot}$ to denote the stop-gradient operator, so that
$\bar{\pi}_\theta(\cdot \mid c, y_{<t}) := \sg{\pi_\theta(\cdot \mid c, y_{<t})}$ denotes
a detached copy of the current model evaluated on context $c$.

\paragraph{Supervised personalization objective.}
The standard way to personalize $\pi_\theta$ is to fine-tune it on the reference response
under teacher forcing, minimizing the token-level negative log-likelihood
\begin{equation}
    \mathcal{L}_{\mathrm{SFT}}(\theta)
    = - \frac{1}{T} \sum_{t=1}^{T}
      \omega_t \log \pi_\theta\!\left(y_t \mid c_{\mathrm{S}}, y_{<t}\right),
    \label{eq:sft}
\end{equation}
where $\omega_t$ is a token-level weight. Setting $\omega_t = 1$ for all $t$ recovers
plain supervised fine-tuning, while recent personalization methods instantiate $\omega_t$
with estimates of how strongly token $y_t$ reflects the user's preferences. Two properties
of Equation~\ref{eq:sft} are central to this paper: the supervision target is the
reference token $y_t$ itself, and it is evaluated on the reference prefix $y_{<t}$ rather
than on a prefix the model would actually produce.

\paragraph{On-policy rollouts.}
Given a deployable context $c_{\mathrm{S}}$, we write
$\hat{y} \sim \pi_\theta(\cdot \mid c_{\mathrm{S}})$ for a response sampled from the
current model, and refer to the resulting prefixes $\hat{y}_{<t}$ as
\emph{on-policy states}. These are the states the model visits at inference time, and they
generally differ from the reference prefixes $y_{<t}$ used in Equation~\ref{eq:sft}.

\paragraph{Divergences between next-token distributions.}
Let $P$ and $Q$ be two distributions over $\mathcal{V}$. We use the generalized
Jensen--Shannon divergence with mixing coefficient $\beta\in[0,1]$,
\begin{equation}
    \JSD_\beta\!\left(P \,\|\, Q\right)
    =
    \beta\,\KL\!\left(P\,\|\,M\right)
    +
    (1-\beta)\,\KL\!\left(Q\,\|\,M\right),
    \qquad
    M=\beta P+(1-\beta)Q,
    \label{eq:jsd}
\end{equation}
which measures the discrepancy between $P$ and $Q$ through their mixture distribution $M$. The coefficient $\beta$ controls the relative contribution of the two KL-to-mixture terms, rather than recovering the forward or reverse KL divergence at the endpoints. In particular, $\beta=0.5$ yields the standard symmetric Jensen--Shannon divergence and treats the two distributions equally. The divergence is finite and bounded for all $P,Q$ when $\beta\in[0,1]$; under natural logarithms, its upper bound is the binary entropy $h(\beta)$ and is at most $\log 2$. Throughout our experiments, we fix $\beta=0.5$ to avoid introducing an additional asymmetry between the reference-revealed teacher and the deployable student. We do not claim that $\beta=0.5$ is optimal; rather, it is adopted as a neutral symmetric setting. A systematic study of alternative mixing coefficients is beyond the scope of this work.
\section{Methodology}
\label{Methodology}
\begin{figure}[t]
    \centering
    \includegraphics[width=0.85\textwidth]{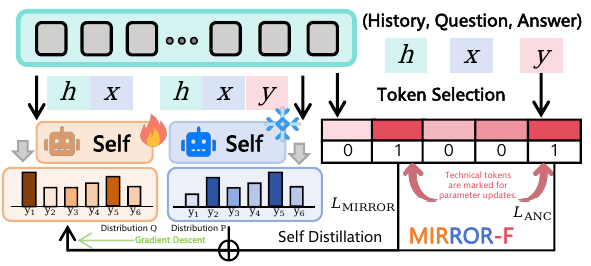}
    \caption{Overview of MIRROR-F. A frozen reference-conditioned teacher ($h,x,y$)
supervises the trainable student ($h,x$) on the student's own rollouts via
$\mathcal{L}_{\text{MIRROR}}$, while $\mathcal{L}_{\text{ANC}}$ anchors only the
selected informative tokens of the reference $y$.}
    \label{fig:methods}
\end{figure}
We present MIRROR, which supervises a model with its own reference-conditioned predictions
over states it visits itself (Section~\ref{sec:mirror}), and MIRROR-F, which adds a sparse
anchoring term whose strength is exposed as a single coefficient
(Section~\ref{sec:mirrorf}).

\subsection{MIRROR: Internalizing Reference-Revealed Reflections}
\label{sec:mirror}

\paragraph{Self-teacher and self-student.}
Rather than treating the reference response $y$ as a string to be reproduced, we treat it
as hindsight information: conditioning on $y$ makes it substantially easier to judge which
continuations suit this user. We therefore evaluate the same model under two contexts, the
reference-revealed $c_{\mathrm{T}} = (h, x, y)$ and the deployable
$c_{\mathrm{S}} = (h, x)$, on a shared prefix $\hat{y}_{<t}$:
\begin{equation}
    p^{\mathrm{T}}_t = \bar{\pi}_\theta\!\left(\cdot \mid c_{\mathrm{T}}, \hat{y}_{<t}\right),
    \qquad
    p^{\mathrm{S}}_t = \pi_\theta\!\left(\cdot \mid c_{\mathrm{S}}, \hat{y}_{<t}\right),
    \label{eq:teacher-student}
\end{equation}
where $\bar{\pi}_\theta = \sg{\pi_\theta}$ is the detached model. Since both distributions
come from the same parameters and the same prefix, their difference isolates the behavioral
shift induced by the reference alone; matching them transfers that shift, rather than the
wording of $y$, into the model, without an external teacher or reward model.

\paragraph{On-policy alignment.}
At each optimization step, a response prefix is sampled from the current
student under stop-gradient sampling,
$\hat{y}\sim\operatorname{sg}[\pi_\theta(\cdot\mid c_{\mathrm{S}})]$,
and the student is supervised on the states visited by this rollout. MIRROR
optimizes the following stopped-gradient surrogate objective:
\begin{equation}
    \widetilde{\mathcal{L}}_{\mathrm{MIRROR}}(\theta)
    =
    \mathbb{E}_{(x,h,y)\sim\mathcal{D}}\,
    \mathbb{E}_{\hat{y}\sim
        \operatorname{sg}[\pi_\theta(\cdot\mid c_{\mathrm{S}})]}
    \left[
      \frac{1}{|\hat{y}|}\sum_{t=1}^{|\hat{y}|}
      \JSD_\beta\!\left(p^{\mathrm{T}}_t
      \,\middle\|\,p^{\mathrm{S}}_t\right)
    \right],
    \label{eq:mirror}
\end{equation}
with $\beta=0.5$. The sampled prefixes and the teacher distribution are
detached during backpropagation, so gradients are propagated only through
the student distribution evaluated on the fixed on-policy states. We use a
mixed divergence because personalization admits multiple valid realizations
for each query: the symmetric objective transfers preference-relevant
probability mass from the reference-revealed teacher while retaining support
for alternative valid continuations.

\paragraph{Why supervision must be on-policy.}
Equation~\ref{eq:mirror} departs from the supervised objective of Equation~\ref{eq:sft} in two respects that jointly determine what the model learns. Its target is a distribution rather than a single reference token, so the model is taught how probability mass should be redistributed instead of which word to emit; and it is evaluated on prefixes the model itself produces, so training and inference operate over the same states. Reweighting $\omega_t$ in Equation~\ref{eq:sft} alters only which reference tokens are emphasized and leaves both properties intact. What MIRROR acquires is therefore a preference-reasoning behavior that transfers to queries for which no reference exists---the only regime that matters at deployment. Check Appendix ~\ref{app:on_policy_gradient} for more details.

\subsection{MIRROR-F: Focal MIRROR with Selective Grounded Anchoring}
\label{sec:mirrorf}

\paragraph{Selective grounded anchoring.}
Distributional alignment governs generation as a whole, whereas professional
usability often depends on a small set of content-bearing tokens, such as
technical terms, named entities, and quantities. MIRROR-F therefore applies
an auxiliary anchoring loss only to selected reference positions. A reference
word is selected when it is \emph{informative} according to the deterministic
rules in Appendix~\ref{app:focal_token_selection} and \emph{grounded}, meaning
that its normalized form occurs as a complete word in the student context
$c_{\mathrm{S}}$. Selection is performed at the word level and expanded to
all subword units of the selected word. Let
$\mathcal{T}_{\mathrm{sel}}$ denote the resulting reference-token indices.
The anchoring loss is
\begin{equation}
\mathcal{L}_{\mathrm{anc}}(\theta)
=
\begin{cases}
-\dfrac{1}{|\mathcal{T}_{\mathrm{sel}}|}
\displaystyle\sum_{t\in\mathcal{T}_{\mathrm{sel}}}
\log \pi_\theta\!\left(y_t\mid c_{\mathrm{S}},y_{<t}\right),
&
|\mathcal{T}_{\mathrm{sel}}|>0,
\\[10pt]
0,
&
|\mathcal{T}_{\mathrm{sel}}|=0,
\end{cases}
\label{eq:anchor}
\end{equation}
where $y_t$ is the $t$-th reference token. Normalization by
$|\mathcal{T}_{\mathrm{sel}}|$ prevents terminology-rich references from
dominating the gradient, while the empty-set convention ensures that the
objective reduces to MIRROR when no token is selected. Complete selection
criteria are provided in Appendix~\ref{app:focal_token_selection}.
\paragraph{Focality coefficient.}
As shown in Figure ~\ref{fig:methods}, because the two terms differ in scale and evolve at different rates, we divide each by a
running average of its own magnitude,
$\tilde{\mathcal{L}}_{k} = \mathcal{L}_{k}/(m_{k} + \epsilon)$ with
$m_{k} \leftarrow \gamma m_{k} + (1-\gamma)\sg{\mathcal{L}_{k}}$ for
$k \in \{\mathrm{MIRROR}, \mathrm{anc}\}$, and combine them through one coefficient
$\alpha \in [0,1]$:
\begin{equation}
    \mathcal{L}_{\mathrm{MIRROR\text{-}F}}(\theta)
    =
    (1-\alpha)\,\tilde{\mathcal{L}}_{\mathrm{MIRROR}}(\theta)
    + \alpha\,\tilde{\mathcal{L}}_{\mathrm{anc}}(\theta).
    \label{eq:mirrorf}
\end{equation}
Equation~\ref{eq:mirrorf} reduces exactly to MIRROR at $\alpha = 0$, and by construction for
any sample whose selected set is empty, so the objective is well defined for all $\alpha$
and all samples. Since personalized style is carried by the distributional term over the
whole trajectory, $\alpha$ does not trade style against content: it sets how sharply
supervision is focused on content-bearing positions on top of an already internalized
behavior. Increasing $\alpha$ thus selects an operating point along the specialization axis,
and moving along it requires no change to the objective, the data, or the pipeline. 

\section{Experiments}
\label{Experiments}
\subsection{Experimental Settings}
\paragraph{Datasets.}
Experiments are conducted on three personalized generation tasks drawn from two
widely used benchmarks for LLM personalization, LaMP~~\citep{lamp} and
LongLaMP~~\citep{longlamp}. Specifically, \textit{Abstract Generation} is taken from
LongLaMP, while \textit{News Headline Generation} (LaMP-4) and \textit{Paper Topic
Generation} (LaMP-5) are taken from LaMP. To ensure that training and evaluation operate over an identical context distribution, we performed context cropping and dataset filtering based on methods from existing papers~\citep{PerCE,nextquill}. For evaluating out-of-distribution (OOD) performance, we further use three benchmarks in Amazon Reviews Dataset~\citep{amazon}, and Music Review, which are not included during training and are reserved exclusively for assessing cross-task transfer and catastrophic forgetting.

\paragraph{Baselines.}
Our methods are compared against three categories of baselines. \textit{(1) Base Model}~\citep{qwen3}: the
backbone LLM is prompted without any personalization and Thinking mode is disabled by default(check Appendix~\ref{app:thinking_ablation} for thinking performance ablation). \textit{(2) Retrieval-based Methods}: RAG~\citep{ragbase}, in which relevant user history is injected into the prompt without updating the model, LatestK~\citep{pplug} selects the most recent $K$ entries by timestamp, and LLM TRSR~\citep{TRSR} compresses the history into a structured
summary via recurrent summarization. \textit{(3) PEFT-based methods}, in which the model is fine-tuned on user-specific data: SFT~\citep{lora} is trained on history-augmented inputs, OPPU~\citep{oppu}
learns a dedicated adapter per user, and PerCE~\citep{PerCE} and NextQuill~\citep{nextquill} additionally reweight the supervision according to token-level preference signals. PerCE and NextQuill are the two most recent state-of-the-art training-based methods and are therefore adopted as the primary points of comparison.

\paragraph{Evaluation Metrics.}
Personalization performance is evaluated from two complementary perspectives. First,for textual similarity against the gold reference, three reference-based metrics are reported: ROUGE-1~\citep{rouge}, METEOR~\citep{meteor}, and BERTScore~\citep{bertscore}, where BERTScore is computed with the \texttt{roberta-large} model. Second, for comprehensive personalization evaluation, we further adopt an LLM-as-judge protocol to assess aspects beyond surface overlap. Textual quality and professionalism are evaluated using two established frameworks: the reference-free G-Eval protocol~\citep{geval} combined with the SummEval dimensions~\citep{summeval}, under which each response is rated on Coherence, Consistency, Fluency, and Relevance; and the reference-based ExPerT framework~\citep{expert}, which decomposes both the reference and the candidate into atomic content and style aspects and reports content- and style-level agreement. The judge model used in all LLM-as-judge evaluations is \texttt{Qwen3-30B-A3B}. For each reference-based evaluation and LLM-as-judge evaluation, we conduct three independent training runs with different random seeds and evaluate each
result under the same protocol. We report the arithmetic mean across the three runs. The judge decoding temperature is fixed to $0$ throughout to ensure reproducible and deterministic scoring for each model output.

\paragraph{Implementation Details} We use Qwen3-1.7B and Qwen3-4B as the backbone LLMs for training and evaluating all methods. For most model training procedures, we employ low-rank adaptation (LoRA)~\citep{lora}. During evaluation, vLLM ~\citep{vllm} is used to accelerate inference. Since different methods use different trained models, we select the best-performing configuration for training and evaluation. For the MIRROR-based methods, each Qwen3-1.7B model is trained for 200 steps, while each Qwen3-4B model is trained for 250 steps. For the main comparison between MIRROR and MIRROR-F, we set the focality coefficient $\alpha$ to 0.5 to ensure a consistent comparison. In the ablation study, we further evaluate $\alpha \in \{0.1, 0.3, 0.5, 0.7, 0.9\}$ to investigate how the focal supervision in MIRROR-F affects personalized styles and syntactic patterns.

\begin{table*}[t]
\centering
\caption{Main results of MIRROR and MIRROR-F on three personalized generation benchmarks \textbf{Bold} numbers denote the best performance, while \underline{underlined} numbers denote the second best.}
\label{tab:main}
\renewcommand{\arraystretch}{1.0}
\setlength{\tabcolsep}{3pt}
\resizebox{\textwidth}{!}{%
\begin{tabular}{llcccccccccc}
\toprule
\multirow{2}{*}{\textbf{Datasets}} & \textbf{Methods ($\rightarrow$)} & \textbf{Base}
  & \multicolumn{3}{c}{\textbf{Retrieval-based Methods}}
  & \multicolumn{4}{c}{\textbf{PEFT-based Methods}}
  & \multicolumn{2}{c}{\textbf{Ours}} \\
\cmidrule(lr){3-3}\cmidrule(lr){4-6}\cmidrule(lr){7-10}\cmidrule(lr){11-12}
 & \textbf{Metrics ($\downarrow$)} & \textbf{Qwen3} & \textbf{RAG} & \textbf{LatestK}
  & \textbf{LLM-TRSR} & \textbf{SFT} & \textbf{OPPU} & \textbf{PerCE} & \textbf{NextQuill}
  & \textbf{MIRROR} & \textbf{MIRROR-F} \\
\midrule
\multicolumn{12}{c}{\textit{Qwen3-1.7B}} \\
\midrule
\multirow{3}{*}{\makecell{\textbf{Abstract}\\\textbf{Generation}}}
 & \textbf{ROUGE-1}   & 0.3717 & 0.3811 & 0.3685 & 0.3758 & 0.3755 & 0.3082 & 0.3670 & 0.3698 & \underline{0.3967} & \textbf{0.4015} \\
 & \textbf{METEOR}    & 0.2125 & 0.2262 & 0.2145 & 0.2311 & 0.2365 & 0.2073 & 0.2240 & 0.2437 & \underline{0.2506} & \textbf{0.2529} \\
 & \textbf{BERTScore} & 0.8577 & 0.8675 & 0.8657 & 0.8656 & 0.8668 & 0.8443 & 0.8621 & \underline{0.8694} & 0.8587 & \textbf{0.8701} \\
\midrule
\multirow{3}{*}{\makecell{\textbf{News Headline}\\\textbf{Generation}}}
 & \textbf{ROUGE-1}   & 0.1422 & 0.1495 & 0.1402 & 0.1387 & 0.1468 & 0.1559 & 0.1527 & 0.1594 & \underline{0.1674} & \textbf{0.1763} \\
 & \textbf{METEOR}    & 0.0976 & 0.1042 & 0.0956 & 0.0957 & 0.0922 & 0.1038 & 0.0976 & \underline{0.1005} & 0.1020 & \textbf{0.1156} \\
 & \textbf{BERTScore} & 0.8465 & 0.8497 & 0.8475 & 0.8461 & 0.8544 & 0.8564 & 0.8559 & \underline{0.8558} & 0.8521 & \textbf{0.8925} \\
\midrule
\multirow{3}{*}{\makecell{\textbf{Paper Topic}\\\textbf{Generation}}}
 & \textbf{ROUGE-1}   & 0.3842 & 0.4052 & 0.3659 & 0.3888 & 0.4104 & 0.4073 & 0.4003 & 0.4095 & \textbf{0.4133} & \underline{0.4368} \\
 & \textbf{METEOR}    & 0.3247 & 0.3516 & 0.3143 & 0.3466 & 0.3383 & 0.3372 & 0.3309 & 0.3350 & \underline{0.3554} & \textbf{0.3721} \\
 & \textbf{BERTScore} & 0.8845 & 0.8887 & 0.8841 & 0.8859 & 0.8923 & 0.8917 & 0.8912 & \underline{0.8917} & 0.8894 & \textbf{0.8922} \\
\midrule
\multicolumn{12}{c}{\textit{Qwen3-4B}} \\
\midrule
\multirow{3}{*}{\makecell{\textbf{Abstract}\\\textbf{Generation}}}
 & \textbf{ROUGE-1}   & 0.3949 & 0.3984 & 0.3884 & 0.3869 & 0.4061 & 0.3697 & 0.3694 & 0.3844 & \underline{0.4027} & \textbf{0.4195} \\
 & \textbf{METEOR}    & 0.2575 & 0.2597 & 0.2496 & 0.2500 & 0.2475 & 0.2225 & 0.2301 & 0.2391 & 0.2659 & \textbf{0.2662} \\
 & \textbf{BERTScore} & 0.8689 & 0.8698 & 0.8680 & 0.8669 & \underline{0.8717} & 0.8635 & 0.8565 & 0.8663 & \textbf{0.8730} & \underline{0.8695} \\
\midrule
\multirow{3}{*}{\makecell{\textbf{News Headline}\\\textbf{Generation}}}
 & \textbf{ROUGE-1}   & 0.1578 & 0.1517 & 0.1462 & 0.1454 & 0.1560 & 0.1593 & 0.1693 & 0.1593 & \underline{0.1691} & \textbf{0.1747} \\
 & \textbf{METEOR}    & 0.1105 & 0.1069 & 0.1014 & 0.1019 & 0.1031 & 0.1022 & 0.1130 & 0.1051 & \underline{0.1142} & \textbf{0.1209} \\
 & \textbf{BERTScore} & 0.8544 & 0.8537 & 0.8519 & 0.8502 & 0.8561 & 0.8561 & 0.8598 & 0.8565 & \underline{0.8567} & \textbf{0.8579} \\
\midrule
\multirow{3}{*}{\makecell{\textbf{Paper Topic}\\\textbf{Generation}}}
 & \textbf{ROUGE-1}   & 0.4118 & 0.4282 & 0.4211 & 0.4224 & 0.4165 & 0.3975 & 0.4257 & \textbf{0.4515} & 0.4376 & \underline{0.4486} \\
 & \textbf{METEOR}    & 0.3954 & 0.3945 & 0.3824 & 0.3937 & 0.3452 & 0.3233 & 0.3548 & 0.3478 & \underline{0.3869} & \textbf{0.3949} \\
 & \textbf{BERTScore} & 0.8917 & 0.8946 & 0.8903 & 0.8893 & 0.8032 & 0.8904 & \underline{0.8942} & 0.8927 & 0.8917 & \textbf{0.8949} \\
\bottomrule
\end{tabular}%
}
\end{table*}
\subsection{Experiment Results}
\paragraph{Reference-based Evaluation} MIRROR and MIRROR-F achieve consistently strong performance on reference-based evaluation metrics, demonstrating that our method possesses superior capabilities for learning personalized styles and sentence structures. As shown in Table~1, the two methods outperform or remain highly competitive with retrieval-based and PEFT-based baselines across all three personalized generation tasks, two model backbones, and the ROUGE-1, METEOR, and BERTScore metrics. The advantage is particularly consistent on ROUGE-1 and METEOR, where MIRROR-F achieves the best results in most settings, including all three tasks with the Qwen3-1.7B backbone, while MIRROR also frequently ranks first or second. The two methods further maintain strong BERTScore performance, indicating that their gains are not limited to lexical overlap but also extend to semantic similarity. These results show that the proposed on-policy internalization paradigm provides a more reliable approach to reference-aligned personalization than conventional retrieval augmentation and token-level preference reweighting.

\paragraph{LLM-based Evaluation} MIRROR and MIRROR-F achieve leading performance under the LLM-based evaluation framework, showing that their advantages extend beyond reference-based similarity to broader aspects of personalized generation quality. Personalized generation is inherently underdetermined, as multiple responses can satisfy the same user preferences and task requirements. Therefore, metrics based on similarity to a single reference may fail to recognize valid alternatives and may not fully reflect content quality or stylistic alignment. As shown in Table~2, MIRROR-F achieves the highest Content-and-Style (C\&S) score comparing to three trained methods across all three tasks, while MIRROR also consistently outperforms or remains competitive with SFT, PerCE, and NextQuill. MIRROR-F further obtains most of the best C\&S scores among the trained methods on 3 benchmarks, indicating that its improvements are stable across tasks and model scales. The consistent gains in both content and style suggest that the proposed methods  internalize preference-relevant information while preserving the content adequacy and stylistic characteristics required for high-quality personalized generation.
\begin{figure}
    \centering
    \includegraphics[width=0.95\textwidth]{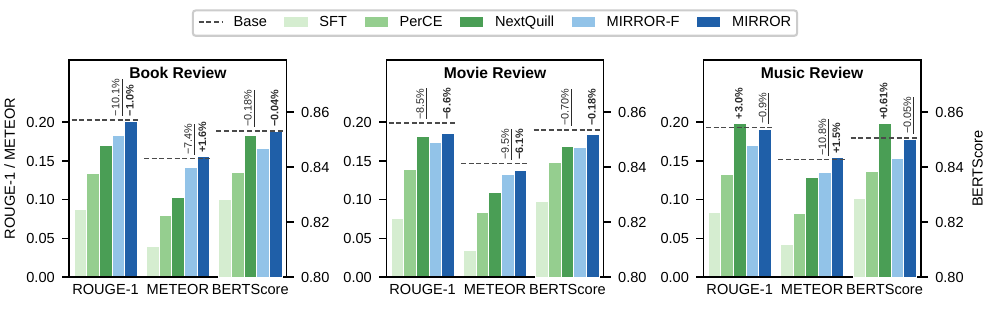}
    \caption{Catastrophic forgetting on held-out OOD personalized generation tasks from the Amazon Review dataset. Performance degradation is measured relative to the corresponding base model under the same evaluation protocol; smaller values indicate better retention. \textbf{Bold} numbers denote the smallest degradation, while \underline{underlined} numbers denote the second-smallest.}
    \label{fig:ood}
\end{figure}
\paragraph{OOD Performance} MIRROR and MIRROR-F exhibit the strongest resistance to catastrophic forgetting on OOD personalized generation tasks, demonstrating superior generalization and scalability compared with existing SFT-based methods. As shown in Figure~\ref{fig:ood}, conventional SFT methods, including PerCE and NextQuill, experience substantial performance degradation on unseen tasks relative to their in-distribution performance, indicating that their strong reliance on reference-specific token-level supervision can lead to over-specialization and impaired transferability. In contrast, MIRROR and MIRROR-F preserve performance much more effectively on the OOD test set and achieve the smallest degradation across the evaluated similarity metrics.

\begin{figure}
    \centering
    \includegraphics[width=0.95\textwidth]{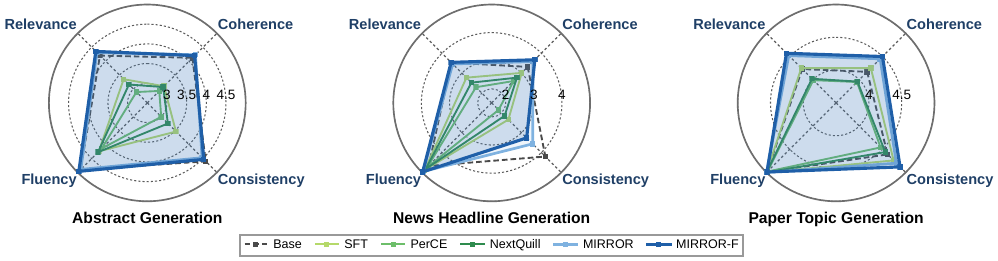}
    \caption{Comparison of model performance scores based on the G-eval framework.}
    \label{fig:geval}
\end{figure}

MIRROR and MIRROR-F consistently outperform existing SFT-based methods across all four dimensions of the third-party text quality evaluation framework. As shown in Figure~\ref{fig:geval}, both methods achieve higher scores on Relevance, Coherence, Consistency, and Fluency than SFT, PerCE, and NextQuill, demonstrating that their improvements extend beyond reference-based similarity to the intrinsic quality of generated responses. MIRROR-F generally obtains the strongest performance, benefiting from its focal supervision of informative and context-grounded content tokens.

\begin{table}
\centering
\caption{ExPert evaluation of personalized generation quality on three
benchmarks. The judge model scores each generated response for \emph{Content},
\emph{Style}, and their combination \emph{C\&S} (Content \& Style); higher is better.
\textbf{Bold} numbers denote the best performance among trained methods, while \underline{underlined} numbers
denote the second best among trained methods.}
\label{tab:judge}
\renewcommand{\arraystretch}{1.0}
\setlength{\tabcolsep}{12pt}
\resizebox{\textwidth}{!}{%
\begin{tabular}{lccccccccc}
\toprule
\textbf{Datasets ($\rightarrow$)}
  & \multicolumn{3}{c}{\textbf{Abstract Generation}}
  & \multicolumn{3}{c}{\textbf{News Headline Generation}}
  & \multicolumn{3}{c}{\textbf{Paper Topic Generation}} \\
\cmidrule(lr){2-4}\cmidrule(lr){5-7}\cmidrule(lr){8-10}
\textbf{Methods ($\downarrow$)}
  & Content & Style & C\&S & Content & Style & C\&S & Content & Style & C\&S \\
\midrule
\multicolumn{10}{c}{\textit{Qwen3-1.7B}} \\
\cmidrule(lr){1-10}
Base  & 0.669 & \underline{8.360} & 7.523 & 0.077 & 4.520 & 2.644 & 0.353 & 8.000 & 5.766 \\
\graydashline
SFT         & 0.563 & 7.990 & 6.808 & \underline{0.080} & 4.730 & \underline{2.765} & 0.417 & 8.360 & 6.267 \\
PerCE       & 0.492 & 7.810 & 6.365 & 0.068 & \underline{4.810} & 2.745 & 0.413 & 8.100 & 6.116 \\
NextQuill   & 0.542 & 8.000 & 6.708 & 0.078 & 4.670 & 2.725 & 0.392 & 8.140 & 6.028 \\
\graydashline
\textbf{MIRROR (Ours)}   & \underline{0.671} & \underline{8.330} & \underline{7.522} & 0.045 & 4.690 & 2.570 & \underline{0.458} & 8.290 & \underline{6.436} \\
\textbf{MIRROR-F (Ours)} & \textbf{0.685} & \textbf{8.390} & \textbf{7.620} & \textbf{0.095} & \textbf{4.890} & \textbf{2.920} & \textbf{0.483} & \textbf{8.440} & \textbf{6.633} \\
\midrule
\multicolumn{10}{c}{\textit{Qwen3-4B}} \\
\cmidrule(lr){1-10}
Base & 0.695 & 8.500 & 7.724 & 0.114 & 5.410 & 3.273 & 0.470 & 8.440 & 6.571 \\
\graydashline
SFT         & 0.600 & 8.080 & 7.042 & 0.078 & 4.970 & 2.876 & 0.438 & 8.310 & 6.345 \\
PerCE       & 0.536 & 7.960 & 6.661 & 0.091 & 4.890 & 2.898 & 0.397 & 8.190 & 6.078 \\
NextQuill   & 0.563 & 8.090 & 6.859 & 0.086 & 4.930 & 2.896 & 0.385 & 8.100 & 5.972 \\
\graydashline
\textbf{MIRROR (Ours)}   & \underline{0.704} & \underline{8.460} & \underline{7.750} & \underline{0.094} & \underline{5.100} & \underline{3.020} & \underline{0.440} & \underline{8.340} & \underline{6.372} \\
\textbf{MIRROR-F (Ours)} & \textbf{0.714} & \textbf{8.520} & \textbf{7.832} & \textbf{0.098} & \textbf{5.270} & \textbf{3.125} & \textbf{0.480} & \textbf{8.350} & \textbf{6.573} \\
\bottomrule
\end{tabular}%
}
\end{table}
\paragraph{Content Quality Performance}
\paragraph{Ablation Study}

\begin{figure}
    \centering
    \includegraphics[width=0.95\textwidth]{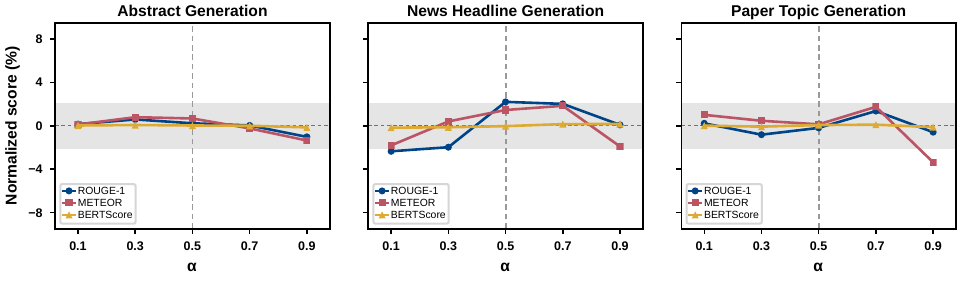}
    \caption{The impact of fluctuations in the value of $\alpha$ on the performance of the model's stylistic and syntactic personalization assessment.}
    \label{fig:ablation}
\end{figure}

The personalization performance of MIRROR-F remains stable across different values of the focality coefficient $\alpha$. As shown in Figure~\ref{fig:ablation}, varying $\alpha$ from $0.1$ to $0.9$ does not lead to a substantial degradation in the evaluated personalization metrics, indicating that the model consistently preserves its ability to generate user-specific styles and sentence structures. This stability suggests that the focal supervision mainly adjusts the emphasis on informative and grounded content tokens without disrupting the underlying preference-relevant behavior learned by MIRROR.

\section{Related Work}
\label{Related Work}

\paragraph{LLM Personalization.}
LLM personalization has been extensively studied across text generation~\citep{rolecde,rolellm}, conversational systems~\citep{coser,Conversation1}, and multimodal applications~\citep{mmpersonal}. Recent benchmarks, including LaMP~\citep{lamp} and LongLaMP~\citep{longlamp}, have provided standardized frameworks for evaluating personalized text generation. Existing methods for LLM personalization can be broadly categorized into three groups: retrieval-based, fine-tuning-based, and user-centric approaches. Retrieval-based methods incorporate user histories into the context through retrieval or augmentation strategies~\citep{ragbase,rag1,rag2,harness,skill,office2,popi}, enabling training-free adaptation but relying on external user information without enhancing the model's internal preference understanding. PEFT
-based methods adapt model parameters using user-specific data, with recent approaches introducing preference-aware supervision reweighting~\citep{PerCE,nextquill}. Although effective under reference-based similarity metrics, they mainly optimize toward reproducing user-specific expressions and suffer from over-adaptation. User-centric methods explicitly model users through dedicated adapters\citep{oppu,OPDpersona}, embeddings, or internal interventions~\citep{pplug,attention,hiddenpref}, but their scalability is limited because of high costs. Existing methods prioritize reference imitation or external preference conditioning, often under-optimizing the intrinsic quality of generated content. In contrast MIRROR introduces a reference-revealed self-distillation training paradigm for personalized generation, while MIRROR-F further focuses optimization on informative content, enhancing the professionalism and overall textual quality of personalized generation.

\paragraph{LLM Self-Distillation.} A recent line of work removes the need for an external, stronger teacher by letting a model conditioned on hindsight information supervise its own weaker self: the teacher
sees privileged signals---environment feedback, demonstrations, or verified reasoning
traces---while the student sees only the query, and training matches their per-token
distributions over the student's own rollouts~\citep{SDPO,OPDpersona,SDFT}. Follow-up work extends
this paradigm to new settings and objectives, including long-trace compaction~\citep{OPSDC},
cross-lingual and long-context reasoning, privacy-constrained training~\citep{OPSDL}, UI generation~\citep{SDPO-UI}and
policy-gradient reformulations~\citep{SDPG}. MIRROR apply on-policy self-distillation to LLM personalization. We further propose a focal plug-in(MIRROR-F) mechanism for personalized generation, which improves the professionalism and textual quality of generated content without compromising personalization performance.
\section{Conclusion}

We introduced MIRROR, an on-policy self-distillation framework that shifts LLM personalization from reference imitation toward preference internalization. We further proposed MIRROR-F, which selectively enhances informative content while preserving user-specific expression. Across multiple tasks, model scales, and evaluation protocols, both methods achieve strong personalization, improved text quality, and less catastrophic forgetting than existing SFT-based approaches, demonstrating the effectiveness of on-policy preference internalization for robust and transferable personalized generation.

\subsection*{Reproducibility statement}

We provide detailed implementation and experimental settings, hyperparameters, training configurations in appendix~\ref{app:training_hyperparameterss}. 

\subsection*{AI use statement}
Generative AI tools were used to assist with code debugging, LaTeX formatting, writing and polishing portions of the manuscript, and literature search for identifying potentially relevant prior work. All AI-assisted outputs, including text, code suggestions, and literature recommendations, were reviewed and verified by the authors before being incorporated into the final work. Generative AI tools were not used to formulate the main research hypotheses, design the proposed methodology or experiments, derive mathematical claims or proofs, generate synthetic data, or interpret the reported experimental results. The authors take full responsibility for the accuracy, validity, and final content of the paper.

\bibliography{reference}
\bibliographystyle{iclr2027_conference}

\appendix
\section{Appendix}
\subsection{More Case Study}
\label{app:more_case_study}

To provide a more detailed qualitative analysis, we present one complete question--answer case for each personalization task. Each case includes the task input, the gold answer, the responses generated by different methods, and a task-specific comparison. The selected examples are drawn from the evaluation set and are intended to illustrate how MIRROR and MIRROR-F balance reference alignment, user-specific expression, and intrinsic content quality.

\subsubsection{Abstract Generation}

\paragraph{Question.}
The input asks the model to generate an abstract for the title ``Learning the Language: The Importance of Studying Written Directions in Designing Navigational Technologies for the Blind.'' The user-provided key information consists of five aspects: independent navigation, text-to-speech directions, language and cues, qualitative analysis, and landmarks and alternate routes.

\paragraph{Analysis.}
The responses exhibit clear differences in both content fidelity and unsupported detail. The base model covers most of the requested concepts, including text-to-speech directions, language, landmarks, and alternate routes, but its discussion remains relatively generic. NextQuill produces a fluent abstract and explicitly mentions landmarks and alternate routes. However, it introduces the unsupported detail that the study involved ``11 blind participants'' and adds a preference for a conversational tone, neither of which is included in the task input or the gold answer.

\begin{table*}
\centering
\small
\caption{Question--answer case for Abstract Generation.}
\label{tab:case_abstract_qa}
\begin{tabular}{p{0.16\textwidth}p{0.78\textwidth}}
\toprule
\textbf{Item} & \textbf{Content} \\
\midrule
\textbf{Question}
&
Generate an abstract for the title ``Learning the Language: The Importance of Studying Written Directions in Designing Navigational Technologies for the Blind'' using the following items: (1) Independent navigation; (2) Text-to-Speech directions; (3) Language and cues; (4) Qualitative analysis; and (5) Landmarks and alternate routes. \\[2mm]

\textbf{Gold answer}
&
Independent navigation is important to individuals who are blind and visually impaired (VI). Researchers have long explored how blind and VI people navigate to inform the design of more useful, accessible wayfinding devices. However, there has been little research on the role language plays in providing effective text-to-speech directions for this population. Through qualitative analysis, this work examines the language and cues expressed in written navigational directions, including how distances are represented, how direction is indicated, and what landmarks are referenced. The analysis further identifies warnings about overshooting a destination, alternative routes that are easier to navigate, and the accessibility of destinations for people with disabilities. \\
\bottomrule
\end{tabular}
\end{table*}

\begin{table*}[t]
\centering
\scriptsize
\caption{Different model answers for the Abstract Generation case.}
\label{tab:case_abstract_answers}
\begin{tabular}{p{0.15\textwidth}p{0.79\textwidth}}
\toprule
\textbf{Method} & \textbf{Answer} \\
\midrule
\textbf{Base}
&
This paper explores the importance of studying written directions in the design of navigational technologies for the blind, focusing on how independent navigation can be enhanced through effective use of text-to-speech directions. We conducted a qualitative analysis of how language and cues are used in written directions to guide users, emphasizing the need for clear, concise, and contextually relevant information. The study also examines the role of landmarks and alternate routes in providing users with multiple options for navigation, ensuring that they can make informed decisions in complex environments. Our findings highlight the significance of incorporating user-centered design principles that consider the unique needs and experiences of blind individuals when developing navigational technologies. \\[2mm]

\textbf{NextQuill}
&
Independent navigation is a challenge for many people who are blind. One promising solution is to use text-to-speech directions to guide them. However, there has been little inquiry into how people who are blind use and interpret directions, and how technologies can be designed to better support them. We conducted a qualitative analysis of 11 blind participants' experiences with and suggestions for text-to-speech directions. We found that users rely on language and cues to interpret directions, and that they had strong preferences for how these directions should be presented. We present two findings that can be used to guide the design of future text-to-speech directions: (1) the use of landmarks and alternate routes can be beneficial to users, and (2) users prefer directions that are presented in a conversational tone. \\[2mm]

\textbf{MIRROR}
&
Independent navigation is a critical skill for individuals who are blind, enabling them to move confidently and safely in their environment. However, the effectiveness of navigational technologies for the blind often hinges on the clarity and usability of the directions provided. This paper explores the importance of studying written directions in the design of navigational technologies for the blind, focusing on how language and cues influence user experience. Through qualitative analysis of user interactions, we examine how text-to-speech directions are perceived and understood by users, highlighting the role of language in conveying information about landmarks and alternate routes. Our findings reveal that the way directions are structured and the cues used can significantly impact the user's ability to navigate independently. We argue that a deeper understanding of language and cues in written directions is essential for designing more effective navigational technologies that support the unique needs of blind users. \\[2mm]

\textbf{MIRROR-F}
&
Independent navigation for the blind is significantly influenced by the clarity and effectiveness of navigational technologies. This study emphasizes the importance of studying written directions in the design of navigational technologies. Through qualitative analysis, we explore how text-to-speech directions, language, and cues impact the user's ability to navigate independently. The study highlights the role of language in conveying information and the necessity of clear, concise cues that aid in understanding the environment. We examine how landmarks and alternate routes are described and how these elements can be integrated into navigational systems to enhance usability. The findings underscore the need for a deep understanding of language in the design process, ensuring that navigational technologies are not only functional but also intuitive and accessible for users with visual impairments. \\
\bottomrule
\end{tabular}
\end{table*}

MIRROR retains the main research focus while avoiding the unsupported participant count. It connects independent navigation with the clarity of directions and explicitly relates language and cues to landmarks and alternative routes. This indicates that the model is not merely reproducing isolated reference phrases, but is reconstructing the central research argument. MIRROR-F further strengthens the content-oriented organization. It explicitly distinguishes text-to-speech directions, language, cues, landmarks, and alternate routes, and connects these elements to the usability and accessibility of navigation systems. The output is also more concise and avoids the speculative participant-level details introduced by NextQuill. The case therefore demonstrates the ability of MIRROR-F to improve content focus while preserving the central semantic structure of the personalized reference.

\subsubsection{News Headline Generation}

\paragraph{Question.}
The input asks the model to generate a headline for a fashion article describing H\&M's updated classics and the appeal of clothing for readers whose ``leggy limbs'' have not appeared in fashion magazines.

\begin{table*}[t]
\centering
\small
\caption{Question--answer case for News Headline Generation.}
\label{tab:case_news_qa}
\begin{tabular}{p{0.16\textwidth}p{0.78\textwidth}}
\toprule
\textbf{Item} & \textbf{Content} \\
\midrule
\textbf{Question}
&
Generate a headline for the following article: ``Um, OK. For those of us whose leggy limbs haven't been immortalized in print, we can turn to H\&M's updated classics, like \ldots'' \\[2mm]

\textbf{Gold answer}
&
H\&M Model Collection Inspired By Joan Smalls, Daphne Groeneveld \& More (PHOTOS) \\
\bottomrule
\end{tabular}
\end{table*}

\begin{table*}[t]
\centering
\small
\caption{Different model answers for the News Headline Generation case.}
\label{tab:case_news_answers}
\begin{tabular}{p{0.15\textwidth}p{0.79\textwidth}}
\toprule
\textbf{Method} & \textbf{Answer} \\
\midrule
\textbf{PerCE}
&
7 Ways To Make Your Pantyhose Look Like A Model (PHOTOS) \\[2mm]

\textbf{NextQuill}
&
Jenna Holmes' '90s Style Is Too Retro For A 2013 Cover (PHOTOS) \\[2mm]

\textbf{MIRROR}
&
H\&M's Updated Classics: Leggy Limbs For Every Body Type (PHOTOS) \\[2mm]

\textbf{MIRROR-F}
&
H\&M's Updated Classics: Leggy Limbs For Every Style (PHOTOS) \\
\bottomrule
\end{tabular}
\end{table*}

\paragraph{Analysis.}
This case highlights the difference between topical relevance and personalized headline generation. PerCE produces a fluent fashion headline, but it changes the article topic from an H\&M collection to general advice about pantyhose. NextQuill preserves the magazine-style ``(PHOTOS)'' suffix, but introduces a different person, a different time period, and a different editorial topic. Both outputs therefore reproduce the broad fashion domain without accurately identifying the article-specific subject.

MIRROR correctly retains ``H\&M'' and ``updated classics,'' which are the most important topical cues in the article. It also preserves the compact tabloid-style structure and the ``(PHOTOS)'' suffix. The phrase ``leggy limbs'' further maintains the distinctive expressive pattern of the input. MIRROR-F retains all of these properties while replacing the narrower ``body type'' expression with the more general ``style.'' This produces a headline that remains faithful to the fashion context and the user's preferred editorial format while being less restrictive in its description of the target audience. The example demonstrates that the proposed methods can preserve both the publication-specific form and the article-specific content, whereas the SFT-based baselines tend to drift toward memorized or superficially related fashion expressions.

\subsubsection{Paper Topic Generation}

\paragraph{Question.}
The input asks the model to generate a title for a paper on converting discrete-time single-input single-output piecewise affine state-space models into equivalent input-output representations. The abstract emphasizes necessary and sufficient conditions, a constructive conversion procedure, the growth in the number of modes and parameters, and numerical examples.

\begin{table*}[t]
\centering
\small
\caption{Question--answer case for Paper Topic Generation.}
\label{tab:case_paper_qa}
\begin{tabular}{p{0.16\textwidth}p{0.78\textwidth}}
\toprule
\textbf{Item} & \textbf{Content} \\
\midrule
\textbf{Question}
&
Generate a title for the following abstract: ``This paper addresses the conversion of discrete-time single input-single output PieceWise Affine (PWA) models from state space to input-output form. Necessary and sufficient conditions are given for a PWA state space model to admit equivalent input-output representations. When an equivalent input-output model exists, a constructive procedure is presented to derive both its parameters and the partition of the regressors domain. It is shown that the number of modes and the number of parameters may grow considerably when converting a PWA state space model into an equivalent input-output representation. Numerical examples highlight the role of the derived necessary and sufficient conditions for input-output realization of PWA state space models.'' \\[2mm]

\textbf{Gold answer}
&
Input-Output Realization Of Piecewise Affine State Space Models \\
\bottomrule
\end{tabular}
\end{table*}

\begin{table*}[t]
\centering
\small
\caption{Different model answers for the Paper Topic Generation case.}
\label{tab:case_paper_answers}
\begin{tabular}{p{0.15\textwidth}p{0.79\textwidth}}
\toprule
\textbf{Method} & \textbf{Answer} \\
\midrule
\textbf{PerCE}
&
On the Input-Output Representation of Piecewise Affine State Space Models \\[2mm]

\textbf{NextQuill}
&
Input-output realization of piecewise affine state space models \\[2mm]

\textbf{MIRROR}
&
Conversion of Discrete-Time Piecewise Affine Models to Input-Output Form \\[2mm]

\textbf{MIRROR-F}
&
Conversion of Piecewise Affine State Space Models to Input-Output Form \\
\bottomrule
\end{tabular}
\end{table*}

\paragraph{Analysis.}
All methods identify the main technical topic, but they differ in how they balance semantic coverage, readability, and title style. PerCE produces a valid title, although ``representation'' is less specific than the ``realization'' terminology used in the reference. NextQuill closely follows the gold title and provides a concise answer, but it mainly reproduces the surface form of the reference.

MIRROR reformulates the title using the operation ``conversion,'' which directly reflects the central procedure described in the abstract. It also preserves the important pair of technical concepts, namely piecewise affine models and input-output form. MIRROR-F produces the most compact formulation among the two proposed methods while retaining the essential technical relation. It removes the unnecessary ``discrete-time'' qualifier for a title-level summary but preserves the central transformation from state-space models to input-output form. The output is therefore concise, technically informative, and consistent with the scholarly title style expressed in the user's history. This example illustrates how MIRROR-F can focus supervision on content-bearing terminology without forcing the model to copy the reference title verbatim.

\paragraph{Overall observation.}
The three cases reveal a consistent distinction between reference imitation and preference internalization. In Abstract Generation, MIRROR and MIRROR-F avoid unsupported details while preserving the research motivation and findings. In News Headline Generation, they maintain the article-specific topic and the user's preferred headline format instead of producing broadly related fashion headlines. In Paper Topic Generation, they preserve the central technical relation while generating concise titles with appropriate scholarly phrasing. MIRROR-F further strengthens the content-bearing portions of the outputs, especially when the task requires the model to identify informative entities, technical concepts, or article-specific themes. These qualitative examples support the view that on-policy reference-revealed self-distillation can improve personalized generation without reducing the factual and practical quality of the generated text.
\subsection{Training Hyperparameters}
\label{app:training_hyperparameterss}

MIRROR and MIRROR-F were trained under the same experimental configuration
for a controlled comparison. Unless otherwise stated, both methods used the
same backbone, training data, context construction procedure, optimizer,
learning rate, batch size, training steps, rollout strategy, and random seed.
The only difference was the training objective: MIRROR used on-policy
distributional alignment alone, whereas MIRROR-F additionally applied the
selective grounded anchoring loss. The shared configuration used for the
comparison is summarized in Table~\ref{tab:training_hyperparameters}.

\begin{table}[t]
\centering
\small
\caption{Training hyperparameters used for the MIRROR and MIRROR-F comparison.}
\label{tab:training_hyperparameters}
\begin{tabular}{lll}
\toprule
\textbf{Category} & \textbf{Hyperparameter} & \textbf{Value} \\
\midrule
\multirow{2}{*}{Model and data}
& Backbone & Qwen3-1.7B/4B \\
& Dataset & LongLaMP,LaMP-4,LaMP-5 \\
\midrule
\multirow{6}{*}{Optimization}
& Optimizer & AdamW \\
& Learning rate & $1\times 10^{-4}$ \\
& Per-device batch size & 1 \\
& Gradient accumulation & 1 \\
& Warm-up ratio & 0.05 \\
& Maximum gradient norm & 1.0 \\
\midrule
\multirow{4}{*}{LoRA}
& Rank $r$ & 16 \\
& Scaling factor $\alpha_{\mathrm{LoRA}}$ & 32 \\
& Dropout & 0.05 \\
& Trainable parameters & LoRA parameters only \\
\midrule
\multirow{5}{*}{On-policy rollout}
& Maximum training steps & 200 (Qwen3-1.7B) / 250 (Qwen3-4B) \\
& Sampling temperature & 1.0 \\
& Top-$p$ & 1.0 \\
& Rollout length ratio & 1.3$\times$ reference length \\
& Minimum / maximum rollout length & 64 / 400 tokens \\
\midrule
\multirow{3}{*}{MIRROR}
& Divergence & Generalized JSD \\
& Mixing coefficient $\beta$ & 0.5 \\
& Anchoring loss & Not used \\
\midrule
\multirow{5}{*}{MIRROR-F}
& Divergence & Generalized JSD \\
& Mixing coefficient $\beta$ & 0.5 \\
& Focal coefficient $\alpha$ & 0.5 \\
\midrule
\multirow{2}{*}{Reproducibility}
& Random seed & 456166,9874,1123 \\
& Thinking mode & Disabled \\
\bottomrule
\end{tabular}
\end{table}

For both methods, the student generated on-policy rollouts using the same
sampling configuration, and the teacher used the same reference-revealed
context. The user history, task questions, context truncation strategy, and
reference-answer budget were also kept identical. Therefore, the comparison
was conducted under one shared training setup, with the selective grounded
anchoring term being the only method-specific addition in MIRROR-F.
\subsection{Rule-Based Information Completeness Analysis}
\label{app:information_completeness}

To complement the LLM-based quality evaluation, we additionally measure
information completeness using a deterministic, rule-based information
recall metric. This metric is designed to quantify whether a model preserves
specific information-bearing words from the reference output, rather than
whether the generated text is generally fluent or semantically coherent.
Therefore, it should be interpreted as a targeted content-coverage measure
and is complementary to the G-Eval professional-quality scores.

For each example $i$, let $g_i$ denote the gold reference, $c_i$ denote the
student input context, and $y_{i,m}$ denote the output generated by model
$m$. We first extract word-like units using the regular expression
\[
\mathcal{W}(x)
=
\left\{
w \mid
w \text{ matches } [\texttt{A-Za-z0-9}][\texttt{A-Za-z0-9'\textbackslash-}]^*
\text{ in } x
\right\}.
\]
This extractor treats sequences containing letters, digits, apostrophes, and
hyphens as word-level units and avoids substring matching. For a word $w$, we
define its alphanumeric core as
\[
\operatorname{core}(w)
=
\operatorname{filter}_{\mathrm{alnum}}(w),
\]
and use lowercase forms for case-insensitive matching.

Following the focal-token selection rules used by MIRROR-F, a word is
considered informative if it contains explicit lexical signals of
information-bearing content. Formally,
\[
\operatorname{Informative}(w)
=
\mathbb{I}\left[
\operatorname{Digit}(w)
\lor
\operatorname{InnerUpper}(w)
\lor
\operatorname{HeadUpper}(w)
\right],
\]
where
\[
\operatorname{Digit}(w)
=
\mathbb{I}\left[
\exists j:\operatorname{core}(w)_j\in\{0,\ldots,9\}
\right],
\]
\[
\operatorname{InnerUpper}(w)
=
\mathbb{I}\left[
\exists j\geq 2:
\operatorname{core}(w)_j \text{ is uppercase}
\right],
\]
and
\[
\operatorname{HeadUpper}(w)
=
\mathbb{I}\left[
|\operatorname{core}(w)|\geq 3
\land
\operatorname{core}(w)_1 \text{ is uppercase}
\land
\operatorname{lower}(\operatorname{core}(w))
\notin\mathcal{S}_{\mathrm{stop}}
\right].
\]
Here, $\mathcal{S}_{\mathrm{stop}}$ is the fixed stopword set used by the
training-time focal selector. Words whose alphanumeric core has fewer than
two characters are not treated as informative.

We further require a candidate reference word to be grounded in the
available student context. Let
\[
\mathcal{C}_i
=
\left\{
\operatorname{lower}(w)
\mid
w\in\mathcal{W}(c_i)
\right\}
\]
denote the set of lowercase word units extracted from the input context.
Then the grounding condition is
\[
\operatorname{Grounded}(w,c_i)
=
\mathbb{I}\left[
\operatorname{lower}(\operatorname{core}(w))
\in \mathcal{C}_i
\right].
\]
Thus, a reference word is counted only when it is both informative and
explicitly recoverable as a complete word from the student input. In
particular, the criterion does not count arbitrary substring matches.

The target set of information-bearing words for example $i$ is therefore
defined as
\[
\mathcal{S}_i
=
\left\{
\operatorname{lower}(\operatorname{core}(w))
\;\middle|\;
w\in\mathcal{W}(g_i),
\;
\operatorname{Informative}(w)=1,
\;
\operatorname{Grounded}(w,c_i)=1
\right\}.
\]
The corresponding output set for model $m$ is
\[
\mathcal{O}_{i,m}
=
\left\{
\operatorname{lower}(w)
\mid
w\in\mathcal{W}(y_{i,m})
\right\}.
\]
Both sets are implemented as sets rather than multisets. Consequently,
repeating the same word multiple times does not increase the score.

For examples with a non-empty target set, the information recall is
computed as
\[
R_{i,m}
=
\frac{
\left|\mathcal{S}_i\cap\mathcal{O}_{i,m}\right|
}{
\left|\mathcal{S}_i\right|
}.
\]
Examples for which $\left|\mathcal{S}_i\right|=0$ are excluded because they
contain no eligible grounded information-bearing word and therefore do not
define a meaningful recall denominator. Let
\[
\mathcal{I}_{m,t}
=
\left\{
i\in\mathcal{D}_t
\mid
|\mathcal{S}_i|>0
\right\}
\]
be the valid examples for model $m$ on task $t$. The reported
information-completeness score is the arithmetic mean of the per-example
recalls:
\[
\operatorname{IC}_{m,t}
=
\frac{1}{|\mathcal{I}_{m,t}|}
\sum_{i\in\mathcal{I}_{m,t}}
R_{i,m}.
\]
This aggregation first computes recall independently for each example and
then averages across examples; it is not obtained by pooling all words from
the dataset into a single global numerator and denominator.

Table~\ref{tab:rule_based_completeness} reports the resulting scores. In the
table, OPD corresponds to MIRROR, while OPD-sftel corresponds to MIRROR-F.
Bold numbers indicate the best performance among trained methods, and
underlined numbers indicate the second-best trained method. The Base model is
included as a non-personalized reference but is excluded from the ranking.

\begin{table*}[t]
\centering
\small
\caption{Rule-based information completeness evaluation. Higher values
indicate better coverage of information-bearing content. Bold numbers denote
the best performance among trained methods, while underlined numbers denote
the second-best. The Base model is reported as a non-personalized reference
and is excluded from the ranking.}
\label{tab:rule_based_completeness}
\begin{tabular}{llcccccc}
\toprule
Backbone & Task
& Base
& ContextSFT
& PerCE
& NextQuill
& MIRROR(Ours)
& MIRROR-F(Ours) \\
\midrule
Qwen3-1.7B
& Abstract
& 0.933
& 0.895
& 0.785
& 0.839
& \underline{0.949}
& \textbf{0.952} \\
& News
& 0.774
& \underline{0.447}
& 0.404
& 0.358
& 0.387
& \textbf{0.631} \\
& Paper
& 0.544
& 0.566
& 0.527
& 0.521
& \underline{0.588}
& \textbf{0.596} \\
\midrule
Qwen3-4B
& Abstract
& 0.951
& 0.919
& 0.854
& 0.819
& \underline{0.957}
& \textbf{0.963} \\
& News
& 0.686
& 0.458
& 0.416
& 0.461
& \textbf{0.676}
& \underline{0.640} \\
& Paper
& 0.645
& 0.564
& 0.557
& 0.543
& \underline{0.615}
& \textbf{0.624} \\
\bottomrule
\end{tabular}
\end{table*}

Several observations emerge from Table~\ref{tab:rule_based_completeness}.
First, MIRROR-F achieves the highest information-completeness score in five
of the six backbone--task combinations. Its advantage is particularly
pronounced on Qwen3-1.7B News, where it reaches $0.631$, compared with
$0.447$ for ContextSFT and $0.387$ for MIRROR. This result suggests that
selective grounded anchoring helps preserve concrete information-bearing
content in a setting where generic supervised adaptation substantially
reduces coverage.

Second, MIRROR and MIRROR-F consistently outperform the other trained
personalization methods on the Abstract and Paper Topic tasks. On Abstract,
the two methods achieve $0.949$ and $0.952$ for Qwen3-1.7B and $0.957$ and
$0.963$ for Qwen3-4B, respectively. The same pattern holds on Paper Topic,
where MIRROR-F obtains $0.596$ and $0.624$ on the two backbones. These results
are consistent with the intended role of selective anchoring: the model is
encouraged to retain grounded, information-bearing content while the
reference-revealed on-policy objective provides broader personalization
guidance.

The only exception is Qwen3-4B News, where MIRROR obtains $0.676$ and
outperforms MIRROR-F at $0.640$. This difference indicates that selective
anchoring is not uniformly beneficial for every task and backbone. In this
case, the unanchored MIRROR objective may provide a better trade-off between
content retention and flexible headline generation. Overall, the rule-based
metric supports the conclusion that MIRROR-family methods generally improve
the preservation of targeted contextual information, while also showing that
the effect of focal anchoring remains task-dependent.

Finally, this metric should not be interpreted as a replacement for
semantic-quality evaluation. It measures exact word-level coverage of a
predefined set of grounded information-bearing terms and does not capture
paraphrases, factual entailment, fluency, coherence, or professional style.
Accordingly, we use it as a transparent diagnostic of information
preservation and interpret it jointly with the G-Eval quality results.
\subsection{Rules for Focal Token Selection}
\label{app:focal_token_selection}

The focal selector in MIRROR-F provides a lightweight mechanism for identifying reference tokens that may carry important, context-accessible information. Rather than applying uniform supervision to the entire reference, additional supervision is concentrated on positions that satisfy two conditions: an informative surface form and lexical presence in the student's input. This design provides an initial approximation to key-information extraction without requiring an external annotator, a named-entity recognition model, or an additional LLM. Importantly, the selector is modular: its rules can be edited to reflect different requirements for text quality while retaining the same on-policy distillation framework.

\paragraph{Selection inputs and outputs.}
Let $c_S$ denote the student's input content, consisting of the retained user history and the current question, and let $y=(y_1,\ldots,y_T)$ denote the tokenized reference. The selector produces a binary mask
\begin{equation}
\mathbf{m}=(m_1,\ldots,m_T),
\qquad m_t\in\{0,1\},
\end{equation}
where $m_t=1$ indicates that reference token $y_t$ receives the additional anchoring supervision. The mask is computed from the reference and the student-side context, not from the teacher's reference-revealed prompt. Consequently, the presence of a word in the reference alone is insufficient for its selection.

\paragraph{Subword grouping.}
Selection is performed over word-like groups before being mapped back to token positions. In the implementation, a new group begins at the first reference token or at a token whose vocabulary representation starts with the whitespace marker \texttt{Ġ} . Consecutive tokens are collected until the next such boundary. For each group $g_j$, the token strings are concatenated, the whitespace markers are removed, and only alphanumeric characters are retained to construct a core string $w_j$.

This grouping prevents a multi-subword term from receiving supervision on only an arbitrary fragment. Once a group is selected, all token positions in that group are marked. Since grouping follows tokenizer whitespace markers rather than a linguistic parser, attached punctuation can also belong to a selected group.

\paragraph{Rule 1: informative surface forms.}
Three inexpensive surface-form tests are used to identify candidate information-bearing groups:
\begin{enumerate}
    \item \textbf{Numeric content.} The core contains at least one digit. This test can identify quantities, years, and alphanumeric identifiers.
    \item \textbf{Internal capitalization.} At least one character after the first character is uppercase. This test can identify abbreviations and mixed-case names, such as \texttt{PWA}, \texttt{DBT}, and \texttt{iPhone}.
    \item \textbf{Initial capitalization.} The core begins with an uppercase character, contains at least three characters, and is not included in a predefined stopword set. This test can identify names and capitalized topic terms.
\end{enumerate}

Groups whose cores contain fewer than two characters are excluded before these tests are applied. In particular, the current rule does not select an isolated single-digit core merely because it is numeric. Formally,
\begin{equation}
I(w)=
\mathbb{I}[|w|\geq 2]\,
\mathbb{I}\!\left[
D(w)\lor U_{\mathrm{inner}}(w)
\lor
\left(
|w|\geq 3
\land U_{\mathrm{first}}(w)
\land \operatorname{lower}(w)\notin\mathcal{B}
\right)
\right],
\label{eq:focal_informative}
\end{equation}
where $D(w)$ indicates the presence of a digit, $U_{\mathrm{inner}}(w)$ indicates an uppercase character after the first position, $U_{\mathrm{first}}(w)$ indicates an uppercase initial character, and $\mathcal{B}$ is the stopword set.

The stopword filter is applied only to the initial-capitalization branch. It suppresses common sentence-initial words such as \texttt{The}, \texttt{This}, and \texttt{We}, while preserving the numeric and internal-capitalization tests. The implementation uses the following set:

\begin{tcolorbox}[
    colback=gray!5,
    colframe=black!45,
    title=Stopwords for the initial-capitalization rule,
    breakable,
    fontupper=\small
]
the, a, an, this, that, these, those, it, its, he, she, they, we,
you, i, in, on, at, of, to, for, and, or, but, if, then, so, as,
is, are, was, were, be, been, being, have, has, had, do, does,
did, will, would, can, could, may, might, must, should, i'm,
it's, there, here, what, when, where, who, why, how, not, no,
yes, with, without, from, by, about, into, over, under, after,
before, my, your, his, her, our, their.
\end{tcolorbox}

These tests are surface-form heuristics rather than semantic entity classification. Their purpose is to obtain a transparent initial set of potentially important terms, not to identify every content-bearing expression.

\paragraph{Rule 2: lexical grounding in the student context.}
An informative group is retained only if its lowercased core is present in the student-side context vocabulary. This vocabulary is constructed using the regular expression
\begin{center}
\verb|[A-Za-z0-9][A-Za-z0-9'\-]*|
\end{center}
and lowercasing the extracted strings. Let $\mathcal{V}(c_S)$ denote the resulting set. The grounding condition is
\begin{equation}
G(w,c_S)
=
\mathbb{I}\!\left[
\operatorname{lower}(w)\in\mathcal{V}(c_S)
\right].
\label{eq:focal_grounding}
\end{equation}

Membership is checked against complete extracted strings rather than arbitrary substrings. For example, the core \texttt{ID} is not considered present merely because the context contains \texttt{video} or \texttt{identify}. The lookup is case-insensitive, so a capitalized reference term can match its lowercase occurrence in the input.

This condition makes the auxiliary target depend on information that is lexically accessible to the student. Here, \emph{grounding} specifically denotes lexical presence; it does not constitute a semantic entailment test. The implementation also uses different normalization procedures for the reference core and the context vocabulary: punctuation is removed from reference cores, whereas internal apostrophes and hyphens are retained in context matches. Thus, expressions such as \texttt{Input-Output} are not automatically treated as identical to the reference core \texttt{InputOutput}. No stemming, synonym matching, or entity linking is applied.

\paragraph{Rule 3: expansion to token positions.}
The final selection criterion is the conjunction of informativeness and lexical grounding:
\begin{equation}
A(w_j,c_S)=I(w_j)G(w_j,c_S).
\end{equation}
The selected token set is then defined as
\begin{equation}
\mathcal{T}_{\mathrm{sel}}
=
\bigcup_{j:A(w_j,c_S)=1}g_j,
\qquad
m_t=\mathbb{I}[t\in\mathcal{T}_{\mathrm{sel}}].
\label{eq:focal_selected_set}
\end{equation}
All subword positions within an accepted group are included. The end-of-sequence token is explicitly excluded from the selective mask.

\paragraph{Illustrative cases.}
The following controlled examples explain the decisions made by the implemented rules. They are constructed illustrations of the selector, rather than additional empirical generation results. The context excerpts shown in each box are treated as the complete context for the corresponding lexical check; no additional occurrences in user history are assumed. Boldface in the reference identifies the word-like groups selected by the rules.

\begin{tcolorbox}[
    colback=gray!5,
    colframe=black!45,
    title={Case 1: Abstract generation---method names and numerical information},
    breakable,
    fontupper=\small
]
\textbf{Student context.}

Generate an abstract for a study of DBT using 120 participants recruited in London.

\medskip
\textbf{Reference illustration.}

This study evaluates \textbf{DBT} with \textbf{120} participants in \textbf{London} and compares outcomes with a Paris cohort.

\medskip
\textbf{Selection decisions.}
\begin{itemize}
    \item \texttt{DBT}: selected because it contains internal uppercase characters and occurs in the context.
    \item \texttt{120}: selected because it contains digits, has at least two characters, and occurs in the context.
    \item \texttt{London}: selected because it is initially capitalized, is not a stopword, and occurs in the context.
    \item \texttt{Paris}: rejected by the grounding condition, despite satisfying the capitalization rule.
    \item \texttt{This}: rejected by the stopword filter.
    \item \texttt{participants}: not selected because its lowercase form does not satisfy an informativeness test.
\end{itemize}
\end{tcolorbox}

This example illustrates how a simple selector can identify several details that are important for scientific reporting: a method abbreviation, a sample size, and a study location. The reference-only location is excluded from the auxiliary target because it is absent from the student's context. The remaining reference positions are not uniformly anchored. Thus, additional supervision is concentrated on selected details without requiring reproduction of the complete reference wording.

\begin{tcolorbox}[
    colback=gray!5,
    colframe=black!45,
    title={Case 2: News headline generation---entities and mixed-case names},
    breakable,
    fontupper=\small
]
\textbf{Student context.}

Generate a headline for the following article: Apple introduced an iPhone update in London.

\medskip
\textbf{Reference illustration.}

\textbf{Apple} Unveils \textbf{iPhone} Update in \textbf{London}

\medskip
\textbf{Selection decisions.}
\begin{itemize}
    \item \texttt{Apple}: selected by initial capitalization and context membership.
    \item \texttt{iPhone}: selected by internal capitalization and context membership, although its first character is lowercase.
    \item \texttt{Update}: selected because the reference form is capitalized and its lowercase form matches \texttt{update} in the article.
    \item \texttt{London}: selected by initial capitalization and context membership.
    \item \texttt{Unveils}: not selected because that exact lexical form is absent from the context.
    \item \texttt{in}: not selected.
\end{itemize}
\end{tcolorbox}

Here, the selected groups recover the organization, product name, event topic, and location. These elements form a useful preliminary description of the news content. The example also shows that the rule is not restricted to formally recognized named entities: title capitalization allows an ordinary topic word such as \texttt{Update} to be selected. This behavior is useful for a lightweight content selector, while making its dependence on the reference's surface form explicit.

\begin{tcolorbox}[
    colback=gray!5,
    colframe=black!45,
    title={Case 3: Paper title generation---technical terminology},
    breakable,
    fontupper=\small
]
\textbf{Student context.}

Generate a title for the following abstract of a paper: This paper studies piecewise affine models and gives conditions for their realization.

\medskip
\textbf{Reference illustration.}

\textbf{Realization} of \textbf{Piecewise} \textbf{Affine} \textbf{Models}

\medskip
\textbf{Selection decisions.}
\begin{itemize}
    \item \texttt{Realization}, \texttt{Piecewise}, \texttt{Affine}, and \texttt{Models}: selected because they are capitalized in the reference and their lowercase forms occur in the abstract.
    \item \texttt{of}: not selected.
    \item If \texttt{Piecewise} is represented by multiple subword tokens, all tokens in its word-like group receive mask value 1.
\end{itemize}
\end{tcolorbox}

The selected positions jointly describe the central technical object and operation of the paper. Although no phrase-level parser is used, adjacent accepted groups can retain a multiword expression such as \emph{piecewise affine models}. This illustrates the intended role of the selector: a small set of explicit rules can already identify useful information-bearing positions, providing a starting point for more specialized annotation policies.

\paragraph{Use in the anchoring objective.}
The mask controls the reference-token loss, not the contents of an additional prompt. In particular, the selected words are not presented to the model as a separate instruction or annotation list. The student is evaluated under its ordinary context and the reference prefix, and the token-level negative log-likelihood is averaged only over selected positions:
\begin{equation}
\mathcal{L}_{\mathrm{anc}}(\theta)
=
-
\frac{
\sum_{t=1}^{T}
m_t\log\pi_{\theta}(y_t\mid c_S,y_{<t})
}{
\max\!\left(1,\sum_{t=1}^{T}m_t\right)
}.
\label{eq:focal_masked_anchoring}
\end{equation}

Normalizing by the number of selected tokens prevents the auxiliary signal from being diluted by the length of the complete reference. Unselected reference tokens remain part of the teacher-forced prefix but do not receive a direct anchoring loss at their own prediction positions. Meanwhile, on-policy distributional alignment continues to operate over the student's generated trajectory.

When no token is selected, the anchoring loss is zero. In the implementation, the anchoring-loss moving average is not updated for that sample, and the effective focal coefficient is set to zero. The update then uses the normalized on-policy alignment term alone. This makes selection optional on a per-sample basis and avoids forcing an arbitrary reference token into the target set.

\paragraph{Customizable selection according to quality requirements.}
The current rules are intentionally simple, interpretable, and inexpensive. Their role is to establish a first-stage extraction mechanism for potentially important information, rather than to prescribe a universal definition of text quality. Different users and applications may require different information to be preserved. For scientific writing, terminology, measurements, and methodological conditions may be prioritized. For news writing, organizations, locations, dates, and event participants may be emphasized. For review generation, product attributes and experience-related expressions may be more relevant.

These requirements can be incorporated by editing the selection policy. Possible extensions include domain-specific terminology lists, explicit handling of units and numerical expressions, phrase-level matching, and user-specified inclusion or exclusion rules. Such changes modify the mask construction while leaving the student prompt and on-policy distillation mechanism unchanged. They also allow lowercase technical expressions to be selected when capitalization is not a useful cue.

\paragraph{Extension to small-LLM-assisted annotation.}
A further direction is to replace or augment the surface-form rules with a small language model that identifies important reference spans under explicit quality requirements. Let $r_u$ encode the user's desired quality criteria. A generalized selector can be written as
\begin{equation}
\mathcal{T}_{\mathrm{sel}}
=
\mathcal{S}_{\phi}(c_S,y,r_u),
\end{equation}
where $\mathcal{S}_{\phi}$ may be a rule system, a compact sequence-labeling model, or a small LLM.

For example, a small LLM could identify technical phrases, distinguish central entities from incidental mentions, and annotate expressions related to numerical accuracy or content completeness. Returned spans could be validated against the reference and mapped to tokenizer positions before training. A separate context-support check could be retained or strengthened to prevent importance annotations from automatically admitting reference-only information. These annotations could be generated offline, so that no additional annotator would be required at inference time.

This learned selector is a prospective extension, not a component of the current implementation. The present rule-based mechanism already supplies the necessary interface: important reference spans are converted into a token mask and consumed by the same anchoring objective. Consequently, improved information identification, user-editable quality criteria, and small-model annotation can be explored within a common framework, providing a concrete foundation for further development of MIRROR-F.


\subsection{Effect of the Anchoring Coefficient $\alpha$ on Professional Quality}
\label{app:alpha_professional_quality}

We further study how the anchoring coefficient $\alpha$ affects the
professional quality of generated outputs. Here, $\alpha$ controls the
relative contribution of the selective grounded anchoring loss in
MIRROR-F. We use the same G-Eval protocol described in
Section~\ref{sec:llm_as_judge}, where the judge model scores each output
along four dimensions, including coherence, consistency, fluency, and
relevance. The overall score is computed as the arithmetic mean of these
four dimensions.

Table~\ref{tab:alpha_quality} reports representative results on the
Scholarly Title task under four different values of $\alpha$. The results
show a non-monotonic relationship between the anchoring strength and
professional quality. The overall score is $4.713$ at $\alpha=0.1$ and
slightly decreases to $4.705$ at $\alpha=0.3$. It then increases to the
highest value of $4.735$ at $\alpha=0.7$, before decreasing substantially
to $4.650$ at $\alpha=0.9$. The consistency score follows a similar
pattern: it reaches its maximum of $4.880$ at $\alpha=0.7$, compared with
$4.810$, $4.830$, and $4.690$ at $\alpha=0.1$, $0.3$, and $0.9$,
respectively. These results indicate that a moderate anchoring coefficient
provides the most favorable trade-off for professional-quality generation.

\begin{table}
\centering
\small
\caption{Representative G-Eval results under different values of
$\alpha$ on the Scholarly Title task. Moderate anchoring achieves the
highest overall professional-quality score, whereas an excessively large
anchoring coefficient leads to performance degradation. Bold numbers denote
the best result in each column.}
\label{tab:alpha_quality}
\begin{tabular}{c|ccccc}
\toprule
$\alpha$
& Coherence
& Consistency
& Fluency
& Relevance
& Overall \\
\midrule
0.1
& 4.480
& 4.810
& 4.980
& \textbf{4.580}
& 4.713 \\
0.3
& 4.440
& 4.830
& 5.000
& 4.550
& 4.705 \\
0.7
& \textbf{4.500}
& \textbf{4.880}
& 4.990
& 4.570
& \textbf{4.735} \\
0.9
& 4.440
& 4.690
& 5.000
& 4.470
& 4.650 \\
\bottomrule
\end{tabular}
\end{table}

This behavior suggests that the selective grounded anchoring loss is most
effective when assigned a moderate weight. When $\alpha$ is relatively
small, the anchoring signal may be insufficient to consistently stabilize
grounded, information-bearing content. Increasing $\alpha$ to a moderate
level improves the balance between preserving relevant content and
maintaining flexible generation, resulting in the highest coherence,
consistency, and overall quality at $\alpha=0.7$. However, when $\alpha$
becomes excessively large, the objective may over-emphasize word-level
anchoring and constrain the model's ability to organize the output
naturally. This can explain the noticeable decline in consistency,
relevance, and overall professional quality at $\alpha=0.9$.

Overall, the results demonstrate that the effect of selective grounded
anchoring is not monotonic. Moderate anchoring is more beneficial than
either weak anchoring or overly strong anchoring, suggesting that the
anchoring coefficient should be carefully controlled to avoid
over-constraining the generation process.

For consistency and fairness, all reported evaluation results in the
main experiments use a fixed anchoring coefficient of $\alpha=0.5$.
\subsection{More Information about LLM-as-Judge Methods}
\label{app:llm_as_judge}

We employ two complementary LLM-as-judge protocols to evaluate personalized generation quality. The first protocol is a reference-free G-Eval procedure based on the four SummEval dimensions, which measures the intrinsic quality of each generated response with respect to the task input. The second protocol is a reference-based, aspect-level evaluation inspired by ExPerT, which measures the alignment between the generated response and the user-specific reference in terms of content and style. The two protocols are complementary: the former evaluates whether an answer is coherent, faithful, fluent, and relevant, whereas the latter evaluates whether the answer preserves the information and expressive characteristics conveyed by the reference.

\subsubsection{Judge Model and General Evaluation Settings}

All LLM-as-judge evaluations are conducted using the locally deployed \texttt{Qwen3-30B-A3B} model. The judge is served with vLLM using tensor parallelism over eight GPUs. To improve reproducibility, the decoding temperature is set to zero, and the judge is instructed not to activate its internal thinking mode. The maximum number of newly generated tokens is set to 512. The same judge model and decoding configuration are used for all compared methods.

For each task, we identify the intersection of sample IDs available for all compared methods. This procedure ensures that every method is evaluated on exactly the same questions. If the number of common samples exceeds the predefined evaluation budget, a fixed random seed is used to select the evaluation subset. Unless otherwise specified, 100 common samples are evaluated for each task. No method-specific sample filtering or method-specific evaluation prompt is applied.

\subsubsection{Reference-Free G-Eval with SummEval Dimensions}

\paragraph{Evaluation objective.}
The reference-free protocol evaluates the intrinsic quality of a candidate response without exposing the gold reference to the judge. For each sample, the judge receives only the original task input and one candidate response generated by one method. The judge does not compare two methods and does not use the gold answer when assigning scores. This design avoids reducing content quality to lexical similarity with a single reference, which is particularly important for personalized generation because multiple responses may satisfy the same user preference.

The evaluation dimensions are adopted from G-Eval and SummEval:

\begin{itemize}
    \item \textbf{Coherence}: whether the response is logically organized and forms a unified answer rather than a collection of loosely related statements.
    \item \textbf{Consistency}: whether the response is factually supported by the source input and avoids hallucinated or unsupported claims.
    \item \textbf{Fluency}: whether the response is grammatical, readable, and appropriately written in terms of spelling, word choice, and punctuation.
    \item \textbf{Relevance}: whether the response preserves the important information from the source input while avoiding redundant or off-topic content.
\end{itemize}

Each dimension is scored from 1 to 5, where 1 denotes the lowest quality and 5 denotes the highest quality. The judge is also asked to provide a short justification before returning the four scores in a fixed machine-readable format.

\paragraph{Task-specific framing.}
The scoring dimensions are identical across tasks. To account for differences in output form, the judge receives one neutral task-form description:

\begin{itemize}
    \item For Abstract Generation, the candidate is described as a research-paper abstract written for a given title.
    \item For News Headline Generation, the candidate is described as a short headline written for a given article.
    \item For Paper Topic Generation, the candidate is described as the title of an academic paper written for a given abstract.
    \item For other personalization tasks, the corresponding task form is specified without changing the scoring criteria.
\end{itemize}

This task-specific framing clarifies the expected genre while keeping the evaluation rubric shared by all methods.

\paragraph{G-Eval scoring prompt.}
The following box presents the prompt template used by the judge. The task description, source input, and candidate output are inserted into the placeholders \texttt{\{TASK\_FORM\}}, \texttt{\{SOURCE\_INPUT\}}, and \texttt{\{CANDIDATE\}}, respectively.

\begin{tcolorbox}[
    colback=gray!5,
    colframe=black!45,
    title=Reference-free G-Eval prompt,
    breakable,
    left=3pt,
    right=3pt,
    top=3pt,
    bottom=3pt
]
\small
\texttt{\textbf{System:} You are a meticulous, impartial NLG evaluation judge.}

\medskip

\texttt{\textbf{User:}}

\texttt{You will be given one CANDIDATE text written for a task. Your job is to rate the candidate on four metrics. Read these instructions carefully and keep them open while reviewing.}

\medskip

\texttt{\{TASK\_FORM\}}

\medskip

\texttt{Evaluation Criteria:}

\texttt{- Coherence (1--5): the collective quality of the response. It should be well-structured and well-organized, not just a heap of related sentences.}

\texttt{- Consistency (1--5): factual alignment between the response and the given source input. A consistent response contains only statements that are entailed by or faithful to the source; penalize hallucinated or unsupported facts.}

\texttt{- Fluency (1--5): the quality of the response in terms of grammar, spelling, word choice, punctuation, and overall readability.}

\texttt{- Relevance (1--5): selection of the important, on-topic content from the source. The response should include salient and pertinent information and exclude redundant or off-topic content.}

\medskip

\texttt{Evaluation Steps:}

\texttt{1. Read the SOURCE INPUT carefully.}

\texttt{2. Read the CANDIDATE and compare it against the source input.}

\texttt{3. Assign an integer score from 1 (worst) to 5 (best) for each metric.}

\medskip

\texttt{SOURCE INPUT:}

\texttt{"""}\\
\texttt{\{SOURCE\_INPUT\}}\\
\texttt{"""}

\medskip

\texttt{CANDIDATE:}

\texttt{"""}\\
\texttt{\{CANDIDATE\}}\\
\texttt{"""}

\medskip

\texttt{Do not use any external reference answer. Give a one or two sentence justification, then end with exactly four lines:}

\texttt{COHERENCE: <1--5>}

\texttt{CONSISTENCY: <1--5>}

\texttt{FLUENCY: <1--5>}

\texttt{RELEVANCE: <1--5>}
\end{tcolorbox}

\paragraph{Per-sample and per-task computation.}
Let $c_{i,m}$, $s_{i,m}$, $f_{i,m}$, and $r_{i,m}$ denote the Coherence, Consistency, Fluency, and Relevance scores assigned to sample $i$ generated by method $m$. The overall quality score for one sample is calculated as the arithmetic mean of the four dimensions:

\begin{equation}
q_{i,m}
=
\frac{
c_{i,m}+s_{i,m}+f_{i,m}+r_{i,m}
}{4}.
\end{equation}

For a task $t$ containing $N_t$ successfully parsed samples, the dimension-level score for method $m$ is calculated as:

\begin{equation}
\overline{d}_{m,t}
=
\frac{1}{N_t}
\sum_{i=1}^{N_t} d_{i,m},
\qquad
d \in \{\mathrm{coh},\mathrm{cons},\mathrm{flu},\mathrm{rel}\}.
\end{equation}

The task-level G-Eval score is computed as the mean of the sample-level overall scores:

\begin{equation}
\overline{q}_{m,t}
=
\frac{1}{N_t}
\sum_{i=1}^{N_t} q_{i,m}.
\end{equation}

If multiple tasks are evaluated, the final model-level score is obtained using an unweighted macro-average over tasks:

\begin{equation}
\overline{q}_{m}
=
\frac{1}{|\mathcal{T}_m|}
\sum_{t\in\mathcal{T}_m}
\overline{q}_{m,t},
\end{equation}

where $\mathcal{T}_m$ denotes the set of tasks for which method $m$ has valid evaluation results. Thus, each task contributes equally to the final score, regardless of the number of samples available for that task. Scores are retained on the original 1--5 scale and are not rescaled to a 0--1 or 0--10 range.

\subsubsection{Reference-Based Aspect-Level Evaluation}

\paragraph{Evaluation objective.}
The reference-based protocol evaluates whether a candidate preserves the information and expressive behavior represented by the gold reference. Unlike the reference-free G-Eval protocol, this evaluation explicitly uses the reference answer. However, it does not compare candidate and reference through token-level overlap. Instead, the judge first decomposes both texts into atomic content and style aspects and then evaluates their alignment.

This aspect-based design is suitable for personalized generation because the candidate may express the same preference in a different wording. A response is therefore rewarded when it preserves the underlying content and style aspects, even when it does not reproduce the exact reference sequence.

\paragraph{Aspect extraction.}
For each reference and candidate, the judge extracts two lists:

\begin{itemize}
    \item \textbf{Content aspects}: atomic factual or semantic points explicitly expressed by the text.
    \item \textbf{Style aspects}: writing characteristics such as tone, structure, formality, voice, and length preference.
\end{itemize}

The extracted aspects are represented as strict JSON objects. The aspect extraction stage is performed before the matching stage. The reference aspects are extracted once for each $(\text{task},\text{sample ID})$ pair and are reused for all methods evaluated on that sample.

\begin{tcolorbox}[
    colback=gray!5,
    colframe=black!45,
    title=Aspect extraction prompt,
    breakable,
    left=3pt,
    right=3pt,
    top=3pt,
    bottom=3pt
]
\small
\texttt{\textbf{System:} You extract atomic aspects from text as strict JSON.}

\medskip

\texttt{\textbf{User:}}

\texttt{Extract the aspects of the following TEXT.}

\texttt{Return STRICT JSON with exactly two keys:}

\texttt{\ \ "content": list of short strings, each an atomic factual/content point the text actually makes;}

\texttt{\ \ "style": list of short strings, each a writing-style attribute (tone, structure, formality, voice, length habit).}

\texttt{Return ONLY the JSON object, no prose.}

\medskip

\texttt{TEXT:}

\texttt{"""}\\
\texttt{\{TEXT\}}\\
\texttt{"""}
\end{tcolorbox}

\paragraph{Aspect matching and scoring.}
After aspect extraction, the judge receives the content and style aspects of the reference and candidate. It then estimates three quantities:

\begin{itemize}
    \item \textbf{Content recall}: the fraction of reference content aspects expressed in the candidate.
    \item \textbf{Content precision}: the fraction of candidate content aspects supported by the reference.
    \item \textbf{Style alignment}: the degree to which the candidate's style matches the style of the reference.
\end{itemize}

Content recall measures whether the candidate preserves the information conveyed by the reference. Content precision penalizes unsupported or ungrounded content introduced by the candidate. Style alignment measures whether the candidate maintains user-specific expressive properties without requiring exact lexical reproduction.

\begin{tcolorbox}[
    colback=gray!5,
    colframe=black!45,
    title=Aspect matching prompt,
    breakable,
    left=3pt,
    right=3pt,
    top=3pt,
    bottom=3pt
]
\small
\texttt{\textbf{System:} You are a meticulous aspect-alignment judge for personalized text.}

\medskip

\texttt{\textbf{User:}}

\texttt{Compare a CANDIDATE against a REFERENCE using their pre-extracted aspects.}

\medskip

\texttt{REFERENCE content aspects: \{REFERENCE\_CONTENT\_ASPECTS\}}

\texttt{REFERENCE style aspects: \{REFERENCE\_STYLE\_ASPECTS\}}

\medskip

\texttt{CANDIDATE content aspects: \{CANDIDATE\_CONTENT\_ASPECTS\}}

\texttt{CANDIDATE style aspects: \{CANDIDATE\_STYLE\_ASPECTS\}}

\medskip

\texttt{Assess alignment:}

\texttt{- CONTENT\_RECALL: fraction 0--1 of REFERENCE content aspects expressed in CANDIDATE;}

\texttt{- CONTENT\_PRECISION: fraction 0--1 of CANDIDATE content aspects supported by REFERENCE;}

\texttt{- STYLE: integer 1--10 for how well CANDIDATE style matches REFERENCE style.}

\medskip

\texttt{End with exactly three lines:}

\texttt{CONTENT\_RECALL: <0--1>}

\texttt{CONTENT\_PRECISION: <0--1>}

\texttt{STYLE: <1--10>}
\end{tcolorbox}

\paragraph{Reference-based score computation.}
For each sample, content precision and content recall are combined using the harmonic mean:

\begin{equation}
F^{\mathrm{content}}_{1,i,m}
=
\frac{
2P_{i,m}R_{i,m}
}{
P_{i,m}+R_{i,m}
},
\end{equation}

where $P_{i,m}$ and $R_{i,m}$ denote content precision and content recall, respectively. If both quantities are zero, the content F1 score is defined as zero.

Because content F1 is defined on the range $[0,1]$ while style is scored on the range $[1,10]$, content F1 is first converted to a ten-point scale. The sample-level overall score is then calculated as:

\begin{equation}
q^{\mathrm{aspect}}_{i,m}
=
\frac{
10F^{\mathrm{content}}_{1,i,m}
+
S_{i,m}
}{2},
\end{equation}

where $S_{i,m}\in[1,10]$ is the style score. The resulting score jointly reflects content preservation and stylistic alignment. Content recall, content precision, content F1, style, and the combined score are all averaged arithmetically over samples within each task:

\begin{equation}
\overline{z}_{m,t}
=
\frac{1}{N_t}
\sum_{i=1}^{N_t} z_{i,m},
\end{equation}

where $z$ can be any of the five aspect-level quantities. As in the G-Eval protocol, the final score across multiple tasks is computed using an unweighted macro-average over task-level scores.

\subsubsection{Relationship between the Two Evaluation Protocols}

The two protocols measure different properties and are therefore reported separately. The reference-free G-Eval score measures the quality of the generated text as a response to the task input:

\begin{equation}
\text{Task Input}
\longrightarrow
\text{Candidate Quality}.
\end{equation}

The reference-based aspect-level score measures how well the candidate preserves reference-revealed content and style:

\begin{equation}
\text{Reference}
\longrightarrow
\text{Content and Style Alignment}.
\end{equation}

Consequently, a high G-Eval score does not require lexical similarity to the gold reference, and a high aspect-level content score does not require the candidate to reproduce the reference verbatim. This distinction is important for personalized generation, where several different outputs may be valid. The combined evaluation framework therefore distinguishes between producing a high-quality response and reproducing the exact wording of a single reference.

In the main experiments, the reference-free G-Eval results are used to assess intrinsic content quality through Coherence, Consistency, Fluency, and Relevance. The reference-based aspect-level results are used to complement the reference-based automatic metrics, such as ROUGE-1, METEOR, and BERTScore, by evaluating semantic content preservation and user-specific style alignment at the aspect level.

\subsection{On-policy Rollout and Gradient Handling}
\label{app:on_policy_gradient}

The objective in Equation~\ref{eq:mirror} is implemented as a
stopped-gradient on-policy distillation objective. At each optimization
step, the current student model first generates one response under the
deployable context $c_{\mathrm{S}}$. The generated token sequence is then
treated as a fixed discrete rollout when the distributional alignment loss
is calculated. Therefore, the implementation does not differentiate through
the discrete sampling operation and does not use a score-function estimator,
REINFORCE term, importance-weighted policy-gradient estimator, or a separate
lagged behavior policy.

More precisely, the rollout is generated as
\begin{equation}
    \hat{y}
    \sim
    \operatorname{sg}\!\left[
        \pi_{\theta}(\cdot\mid c_{\mathrm{S}})
    \right],
\end{equation}
where $\operatorname{sg}[\cdot]$ denotes stop-gradient. For each sampled
prefix $\hat{y}_{<t}$, the student distribution is computed as
\begin{equation}
    p_t^{\mathrm{S}}
    =
    \pi_{\theta}
    \left(
        \cdot\mid c_{\mathrm{S}},\hat{y}_{<t}
    \right),
\end{equation}
while the reference-revealed teacher distribution is detached:
\begin{equation}
    p_t^{\mathrm{T}}
    =
    \operatorname{sg}\!\left[
        \pi_{\theta}
        \left(
            \cdot\mid c_{\mathrm{T}},\hat{y}_{<t}
        \right)
    \right].
\end{equation}
Consequently, the gradient is propagated through $p_t^{\mathrm{S}}$ only.
The sampled tokens, the rollout probabilities, and $p_t^{\mathrm{T}}$ do not
receive gradients. The student model is updated after the loss computation,
and a new rollout is generated from the updated student at the next
optimization step.

The rollout uses stochastic sampling with temperature $1.0$ and
$\mathrm{top}\text{-}p=1.0$. The maximum generation length is determined on
a per-example basis from the reference length and the task-specific cap.
Generation terminates when either the EOS token is produced or the maximum
length is reached. If EOS is produced before the length limit, the EOS token
is retained in the aligned completion and tokens after EOS are discarded.
If the length limit is reached without EOS, the capped completion is used and
the event is recorded for diagnostic purposes. Samples with fewer than two
completion tokens are skipped because they do not provide a meaningful
sequence-level alignment signal.

The implementation therefore optimizes a surrogate objective rather than the
full derivative of an expectation whose sampling distribution depends on
$\theta$. In particular, the gradient does not contain the additional
score-function term
\begin{equation}
    \mathbb{E}_{\hat{y}\sim\pi_\theta}
    \left[
        \mathcal{L}(\hat{y},\theta)
        \nabla_\theta\log\pi_\theta(\hat{y}\mid c_{\mathrm{S}})
    \right].
\end{equation}
This choice avoids the high variance associated with policy-gradient
estimation and directly trains the student to match the teacher distribution
on states that the deployable student actually visits. The term
``on-policy'' therefore refers to the state distribution used to construct
the training prefixes, whereas the distributional alignment itself is
optimized by ordinary backpropagation on the detached sampled trajectories.

\subsection{Other Prompts}
\label{app:other_prompts}

This section describes the prompts used for model inference, user-history conditioning, and MIRROR training. The prompts used for LLM-as-judge evaluation are described separately in Appendix~\ref{app:llm_as_judge}. All prompts are converted into the backbone model's native chat format using the tokenizer's chat template. Unless otherwise stated, the system message is shared across training and evaluation.

\subsubsection{Common Generation Prompt}

For all generation-based evaluations, the model receives a system instruction followed by a user message. The system instruction constrains the model to output only the requested text and prevents additional explanations or formatting from being included in the prediction.

\begin{tcolorbox}[
    colback=gray!5,
    colframe=black!45,
    title=Common generation prompt,
    breakable,
    left=3pt,
    right=3pt,
    top=3pt,
    bottom=3pt
]
\small
\texttt{\textbf{System:}}

\texttt{You are a helpful assistant. Write only the requested text itself, with no preamble, no explanation, and no surrounding quotation marks.}

\medskip

\texttt{\textbf{User:}}

\texttt{\{USER\_CONDITIONING\}}

\texttt{\{TASK\_QUESTION\}}
\end{tcolorbox}

Here, \texttt{\{TASK\_QUESTION\}} denotes the original task instruction and input, while \texttt{\{USER\_CONDITIONING\}} is either empty, a retrieved user profile, or the full user history, depending on the evaluation protocol. The model is always evaluated using the same system instruction, regardless of whether it is a base model, an SFT-based model, or a MIRROR-based model.

For example, when no user history is supplied, the user message consists only of the task question:

\begin{tcolorbox}[
    colback=gray!5,
    colframe=black!45,
    title=Non-personalized generation prompt,
    breakable,
    left=3pt,
    right=3pt,
    top=3pt,
    bottom=3pt
]
\small
\texttt{\textbf{System:}}

\texttt{You are a helpful assistant. Write only the requested text itself, with no preamble, no explanation, and no surrounding quotation marks.}

\medskip

\texttt{\textbf{User:}}

\texttt{\{TASK\_QUESTION\}}
\end{tcolorbox}

The tokenizer-specific chat template is applied after the system and user messages have been constructed. For Qwen3 models, the evaluation configuration disables thinking mode by default. Thus, the model directly generates the requested answer without producing an additional \texttt{<think>} segment.

\subsubsection{Task Question Templates}

The original task question is preserved during evaluation. We use the following task-specific forms.

\paragraph{Abstract Generation.}
For Abstract Generation, the model is asked to generate an abstract conditioned on a paper title and a set of key points:

\begin{tcolorbox}[
    colback=gray!5,
    colframe=black!45,
    title=Abstract Generation question template,
    breakable,
    left=3pt,
    right=3pt,
    top=3pt,
    bottom=3pt
]
\small
\texttt{Generate an abstract for the title ``\{TITLE\}'' using the following items:}

\texttt{1. \{KEY\_POINT\_1\}}

\texttt{2. \{KEY\_POINT\_2\}}

\texttt{\ldots}

\texttt{\{K\}. \{KEY\_POINT\_K\}}
\end{tcolorbox}

The output is expected to be a self-contained research-paper abstract. The model is not explicitly asked to copy the reference abstract. Instead, the key points and the user history provide the information and stylistic context needed to generate a new abstract.

\paragraph{News Headline Generation.}
For News Headline Generation, the model receives a news article and is asked to generate a headline:

\begin{tcolorbox}[
    colback=gray!5,
    colframe=black!45,
    title=News Headline Generation question template,
    breakable,
    left=3pt,
    right=3pt,
    top=3pt,
    bottom=3pt
]
\small
\texttt{Generate a headline for the following article:}

\texttt{\{ARTICLE\_TEXT\}}
\end{tcolorbox}

The generated answer is expected to be a short headline rather than a paragraph. The task therefore evaluates whether the model can identify the article-specific topic while preserving the user's preferred headline format, vocabulary, and editorial style.

\paragraph{Paper Topic Generation.}
For Paper Topic Generation, the model receives an academic abstract and is asked to produce a paper title:

\begin{tcolorbox}[
    colback=gray!5,
    colframe=black!45,
    title=Paper Topic Generation question template,
    breakable,
    left=3pt,
    right=3pt,
    top=3pt,
    bottom=3pt
]
\small
\texttt{Generate a title for the following abstract of a paper:}

\texttt{\{PAPER\_ABSTRACT\}}
\end{tcolorbox}

The output is expected to be a concise academic title. The model should preserve the central technical concepts of the abstract while matching the user's preferred terminology and title structure.

\paragraph{Other LongLaMP tasks.}
For product review generation, the question specifies the product and the requested review context. For topic writing, the question provides the title or prompt for a Reddit-style post body. The same common generation prompt is used, while only the task question changes.

\subsubsection{User-History Conditioning}

Personalization is introduced by appending user-specific historical examples before the current task question. The historical examples are formatted as demonstrations that associate an input with the corresponding user-written output. The formatting depends on the task because different tasks contain different input--output structures.

\paragraph{Full-history conditioning.}
In the full-history setting, historical examples are retained in their original temporal order. When the complete history exceeds the input budget, the most recent entries are preferentially retained. The current question is always preserved in full whenever possible. The resulting user message is:

\begin{tcolorbox}[
    colback=gray!5,
    colframe=black!45,
    title=Full-history conditioning template,
    breakable,
    left=3pt,
    right=3pt,
    top=3pt,
    bottom=3pt
]
\small
\texttt{\{HISTORY\_EXAMPLE\_1\}, and}

\texttt{\{HISTORY\_EXAMPLE\_2\}, and}

\texttt{\ldots, and}

\texttt{\{HISTORY\_EXAMPLE\_M\}. Following the given patterns, \{TASK\_QUESTION\}}
\end{tcolorbox}

The phrase ``Following the given patterns'' indicates that the preceding examples describe the user's writing behavior. It does not instruct the model to copy any particular historical answer verbatim.

\paragraph{Task-specific history formatting.}
For Abstract Generation and Paper Topic Generation, each historical item is rendered as:

\begin{tcolorbox}[
    colback=gray!5,
    colframe=black!45,
    title=Academic history example,
    breakable,
    left=3pt,
    right=3pt,
    top=3pt,
    bottom=3pt
]
\small
\texttt{``\{TITLE\}'' is a title for the abstract ``\{ABSTRACT\}''}
\end{tcolorbox}

For News Headline Generation, each historical item is rendered as:

\begin{tcolorbox}[
    colback=gray!5,
    colframe=black!45,
    title=News history example,
    breakable,
    left=3pt,
    right=3pt,
    top=3pt,
    bottom=3pt
]
\small
\texttt{``\{HEADLINE\}'' is a headline for the article ``\{ARTICLE\_TEXT\}''}
\end{tcolorbox}

For product review generation, a historical item contains the review summary, rating, product description, and user-written review:

\begin{tcolorbox}[
    colback=gray!5,
    colframe=black!45,
    title=Product-review history example,
    breakable,
    left=3pt,
    right=3pt,
    top=3pt,
    bottom=3pt
]
\small
\texttt{the review with summary ``\{SUMMARY\}'' and rating \{RATING\} for the product ``\{DESCRIPTION\}'' is ``\{REVIEW\}''}
\end{tcolorbox}

For Amazon review tasks, the historical item is rendered as:

\begin{tcolorbox}[
    colback=gray!5,
    colframe=black!45,
    title=Amazon-review history example,
    breakable,
    left=3pt,
    right=3pt,
    top=3pt,
    bottom=3pt
]
\small
\texttt{the review with rating \{RATING\} and title ``\{REVIEW\_TITLE\}'' for the item ``\{ITEM\_TITLE\}'' (\{ITEM\_DESCRIPTION\}) is ``\{REVIEW\_TEXT\}''}
\end{tcolorbox}

For topic writing, historical examples are rendered as:

\begin{tcolorbox}[
    colback=gray!5,
    colframe=black!45,
    title=Topic-writing history example,
    breakable,
    left=3pt,
    right=3pt,
    top=3pt,
    bottom=3pt
]
\small
\texttt{the reddit post with summary ``\{SUMMARY\}'' is ``\{POST\_BODY\}''}
\end{tcolorbox}

These task-specific templates preserve the semantic relation between a historical input and the corresponding user-written output. Consequently, the model can infer both the user's preferred content selection and the user's preferred expression pattern.

\subsubsection{BM25 Retrieval Prompt}

For retrieval-based baselines, the complete user history is not appended to every query. Instead, BM25 is applied to the user profile, and the two most relevant historical examples are selected. The retrieval corpus is constructed from both the input and output fields of each historical item. For example, the corpus for Abstract Generation contains the paper title and abstract, while the corpus for News Headline Generation contains the headline and article text.

The selected examples are rendered using the task-specific history templates above and concatenated with the current task question:

\begin{tcolorbox}[
    colback=gray!5,
    colframe=black!45,
    title=BM25 retrieval-conditioned prompt,
    breakable,
    left=3pt,
    right=3pt,
    top=3pt,
    bottom=3pt
]
\small
\texttt{\{RETRIEVED\_EXAMPLE\_1\}, and \{RETRIEVED\_EXAMPLE\_2\}.}

\texttt{Following the given patterns, \{TASK\_QUESTION\}}
\end{tcolorbox}

The retrieval operation is performed independently for each test question. It affects only the prompt provided to the model and does not modify the model parameters. The gold answer is never used to construct the retrieval prompt.

\subsubsection{MIRROR Student Prompt}

The MIRROR student is trained to operate under the same information available during deployment. Therefore, the student receives the user history and the current task question, but not the gold reference answer. The student prompt is identical to the full-history inference prompt:

\begin{tcolorbox}[
    colback=gray!5,
    colframe=black!45,
    title=MIRROR student prompt,
    breakable,
    left=3pt,
    right=3pt,
    top=3pt,
    bottom=3pt
]
\small
\texttt{\textbf{System:}}

\texttt{You are a helpful assistant. Write only the requested text itself, with no preamble, no explanation, and no surrounding quotation marks.}

\medskip

\texttt{\textbf{User:}}

\texttt{\{FULL\_USER\_HISTORY\}}

\texttt{Following the given patterns, \{TASK\_QUESTION\}}
\end{tcolorbox}

During training, the current student model samples an on-policy response from this prompt. Let $\hat{y}$ denote the sampled response and $\hat{y}_{<t}$ denote its prefix at step $t$. The student distribution is evaluated as

\begin{equation}
p_t
=
\pi_{\theta}
\left(
\cdot \mid
c_S,\hat{y}_{<t}
\right),
\end{equation}

where $c_S$ contains only the information available to the student at deployment time.

\subsubsection{MIRROR Teacher Prompt}

The MIRROR teacher receives the same user history and task question as the student, together with the reference answer. The reference is treated as privileged hindsight information used only during training. It is not provided to the deployed model.

The teacher prompt is constructed by placing the reference answer after the task question and adding an instruction that asks the teacher to solve the task independently after understanding the reference:

\begin{tcolorbox}[
    colback=gray!5,
    colframe=black!45,
    title=MIRROR teacher prompt,
    breakable,
    left=3pt,
    right=3pt,
    top=3pt,
    bottom=3pt
]
\small
\texttt{\textbf{System:}}

\texttt{You are a helpful assistant. Write only the requested text itself, with no preamble, no explanation, and no surrounding quotation marks.}

\medskip

\texttt{\textbf{User:}}

\texttt{\{FULL\_USER\_HISTORY\}}

\texttt{\{TASK\_QUESTION\}}

\texttt{Here is a reference solution:}

\texttt{\{GOLD\_REFERENCE\}}

\texttt{After understanding the reference solution, please try to solve this problem using your own approach below:}
\end{tcolorbox}

The teacher is evaluated on the student's sampled prefix rather than on the reference prefix. Specifically, the teacher distribution is computed as

\begin{equation}
q_t
=
\overline{\pi}_{\theta}
\left(
\cdot \mid
c_T,\hat{y}_{<t}
\right),
\end{equation}

where $c_T$ is the reference-revealed context and $\overline{\pi}_{\theta}$ denotes a detached copy of the current model. The reference answer therefore changes the teacher's distribution, but it is not used as a token-level target sequence.

The student and teacher prompts are intentionally different in information content:

\begin{equation}
c_S
=
(\text{user history},\text{task question}),
\end{equation}

\begin{equation}
c_T
=
(\text{user history},\text{task question},
\text{reference answer},\text{reference hint}).
\end{equation}

The teacher and student are nevertheless evaluated on the same on-policy prefix $\hat{y}_{<t}$. This construction transfers the behavior induced by the reference into states that the student actually visits during inference.

\subsubsection{MIRROR-F Teacher Prompt and Focal Anchoring}

MIRROR-F uses the same student and teacher prompts as MIRROR. The difference lies in the training objective rather than in the basic prompt structure. In addition to aligning the student and teacher distributions over the student's on-policy trajectory, MIRROR-F applies a selective anchoring loss to informative reference tokens.

The selected token set contains content-bearing and context-grounded tokens. In particular, the selection procedure favors terminology, named entities, quantities, and other tokens that carry task-specific information, while excluding purely functional or stylistic tokens. The selected word-level positions are expanded to their complete subword spans.

The focal anchoring prompt therefore does not expose any additional information to the deployed model. It only determines which reference tokens receive the auxiliary training signal:

\begin{tcolorbox}[
    colback=gray!5,
    colframe=black!45,
    title=MIRROR-F selective anchoring template,
    breakable,
    left=3pt,
    right=3pt,
    top=3pt,
    bottom=3pt
]
\small
\texttt{Reference solution:}

\texttt{\{GOLD\_REFERENCE\}}

\medskip

\texttt{Selected informative and context-grounded positions:}

\texttt{\{SELECTED\_REFERENCE\_TOKENS\}}
\end{tcolorbox}

The selective anchoring term is applied only during training. At inference time, MIRROR-F receives exactly the same deployable prompt as MIRROR:

\begin{equation}
\text{Inference input}_{\mathrm{MIRROR-F}}
=
\text{Inference input}_{\mathrm{MIRROR}}.
\end{equation}

This separation is essential. The gold reference and the selected reference tokens are privileged training information and are never supplied to the deployed model.

\subsubsection{User Profile Summary Prompt}

For the summary-based ablation, a compact user profile is generated from the user's historical texts. The profile summary is generated without access to the current gold answer. It is therefore intended to compress writing preferences rather than reveal the target output.

The summary prompt is:

\begin{tcolorbox}[
    colback=gray!5,
    colframe=black!45,
    title=User profile summary prompt,
    breakable,
    left=3pt,
    right=3pt,
    top=3pt,
    bottom=3pt
]
\small
\texttt{\textbf{System:}}

\texttt{You are a helpful assistant. Write only the requested text itself, with no preamble, no explanation, and no surrounding quotation marks.}

\medskip

\texttt{\textbf{User:}}

\texttt{Here is the writing history of a single user (their past examples):}

\texttt{\{USER\_HISTORY\}}

\medskip

\texttt{Based only on the writing history above, write a complete user profile report describing the user's preferences for content selection, organization, tone, terminology, sentence structure, and level of detail for \{TASK\_DESCRIPTION\}. Be specific and evidence-based. Do not quote whole sentences from the history and do not mention that you were given a history.}
\end{tcolorbox}

The generated summary is inserted only into the privileged teacher context in the summary-based variant. The student still receives the original user history and task question. The summary is truncated to a fixed token budget before being inserted into the teacher prompt.

\subsection{Thinking-Mode Ablation}
\label{app:thinking_ablation}

We further examine whether enabling the model's internal thinking mode
substantially changes personalized generation performance. Table~\ref{tab:thinking_ablation}
reports the results of the same Qwen3 backbones under thinking-on and
thinking-off configurations. The reported values are macro-averaged across
the three LongLaMP tasks, while the task-level results are provided to show
the variation across generation settings.

\begin{table}
\centering
\small
\caption{Thinking-mode ablation on LongLaMP. Values are macro-averaged across
Abstract Generation, Product Review, and Topic Writing. $\Delta$ denotes the
change from thinking off to thinking on; positive values indicate improvement.}
\label{tab:thinking_ablation}
\begin{tabular}{lccc|ccc}
\toprule
& \multicolumn{3}{c|}{Qwen3-1.7B}
& \multicolumn{3}{c}{Qwen3-4B} \\
\textbf{Mode}
& ROUGE-1 & METEOR & BERTScore
& ROUGE-1 & METEOR & BERTScore \\
\midrule
Thinking off
& 0.2661 & 0.1437 & 0.8378
& 0.3334 & 0.1985 & 0.8467 \\
Thinking on
& 0.2908 & 0.1616 & 0.8351
& 0.3220 & 0.1847 & 0.8340 \\
$\Delta$
& +0.0247 & +0.0179 & $-0.0027$
& $-0.0114$ & $-0.0138$ & $-0.0127$ \\
\bottomrule
\end{tabular}
\end{table}

The results show that enabling thinking does not induce a consistent
performance trend in personalized generation. For Qwen3-1.7B, thinking mode
improves the macro-averaged ROUGE-1 and METEOR scores, while producing a
slightly lower BERTScore. For Qwen3-4B, thinking mode decreases all three
macro-averaged metrics, although it improves ROUGE-1 and METEOR on the Topic
Writing task. These results indicate that the effect of thinking is
task- and scale-dependent rather than uniformly beneficial or harmful.
Overall, the main performance trend of personalized generation remains
largely preserved across the two thinking configurations.

This observation suggests that personalized generation and long-horizon
reasoning should be regarded as related but distinct capability dimensions.
Thinking mode is designed to allocate additional computation to intermediate
reasoning, whereas personalization requires the model to infer and apply
user-specific preferences in content selection, organization, terminology,
and expression. Improvements in complex reasoning ability alone therefore do
not necessarily translate into improvements in personalized generation.
Likewise, the effectiveness of RL- or OPD-based training for long-horizon
problem solving should not be assumed to imply a corresponding gain in
personalization. The two capabilities may rely on different forms of
internal representation and different training signals.

Accordingly, these results motivate the study of LLM personalization as an
independent research problem rather than treating it as an automatic
consequence of general reasoning improvement. Methods that explicitly
internalize user preferences, such as MIRROR and MIRROR-F, address this
separate dimension by introducing preference-revealed supervision and
personalization-specific objectives. Further investigation is needed to
understand how reasoning ability and personalization ability interact and
whether they can be improved jointly without assuming that progress in one
dimension automatically transfers to the other.

\end{document}